\documentclass[10pt]{asme2ej_HAL}
\usepackage{lmodern}
\usepackage{graphicx} 
\usepackage{mathrsfs}

\usepackage{epsfig} 
\usepackage{amsmath} 
\usepackage{amssymb}  
\usepackage{hhline} 
\usepackage{subfig} 
\usepackage{multirow}
\usepackage{bm}
\usepackage{leftidx}
\usepackage{enumerate}
\usepackage{caption}
\usepackage{epstopdf}
\usepackage{textcomp}
\usepackage{float}
\usepackage{microtype}
\usepackage{nomencl}
\usepackage{graphicx}
\usepackage{xcolor}
\usepackage{colortbl}
\usepackage[overload]{empheq}
\usepackage[left,modulo]{lineno}

\usepackage{ulem}

\newcommand{\beq}{\begin{equation}}
\newcommand{\eeq}{\end{equation}}
\newcommand{\beqnn}{\begin{equation*}}
\newcommand{\eeqnn}{\end{equation*}}
\newcommand{\beqa}{\begin{eqnarray}}
\newcommand{\eeqa}{\end{eqnarray}}
\newcommand{\beqann}{\begin{eqnarray*}}
\newcommand{\eeqann}{\end{eqnarray*}}
\newcommand{\bseq}{\begin{subequations}}
\newcommand{\eseq}{\end{subequations}}
\newcommand{\bes}{\begin{split}}
\newcommand{\ees}{\end{split}}

\newlength{\largeurargument}
\newlength{\hauteurargument}

\newcommand{\Rbb}{\mathbb{R}}

\title{Input-Shaping for Feed-Forward Control of Cable-Driven Parallel Robots}

\author{Sana Baklouti
	\affiliation{
		University of Nantes,\\
	IUT Nantes, LS2N UMR CNRS 6004,\\
		2 Avenue du Professeur Jean Rouxel,\\
		44475 Carquefou, France \\
		Email: sana.baklouti@ls2n.fr}	
}

\author{Eric Courteille
	\affiliation{
		University of Rennes,\\
		INSA Rennes, LGCGM-EA 3913,\\
		20, avenue des Buttes de C\"{o}esmes,\\
		35043 Rennes Cedex, France\\
		Email: eric.courteille@insa-rennes.fr
	}	
}

\author{Philippe Lemoine
	\affiliation{
		Centrale Nantes, \\
		LS2N UMR CNRS 6004,\\
		1, rue de la No\"e,\\
		44321 Nantes Cedex 03, France\\
		Email: Philippe.lemoine@ls2n.fr
	}	
}

\author{St\'{e}phane Caro\thanks{Corresponding author}
	\affiliation{
		CNRS, \\
		LS2N UMR CNRS 6004,\\
		1, rue de la No\"e,\\
		44321 Nantes Cedex 03, France\\
		Email: stephane.caro@ls2n.fr
	}	
}

\begin{document}

\maketitle    
\begin{abstract}
{\it This paper deals with the use of input-shaping filters in conjunction with a feed-forward control of Cable-Driven Parallel Robots (CDPRs), while integrating cable tension calculation to satisfy positive cable tensions along the prescribed trajectory of the moving-platform. This method aims to attenuate the oscillatory motions of the moving-platform. Thus, the input signal is modified to make it self-cancel residual vibrations. The effectiveness, in terms of moving-platform oscillation attenuation, of the proposed closed-loop control method combined with shaping inputs is experimentally studied on a suspended and non-redundant CDPR prototype. This confirms residual vibration reduction improvement with respect to the unshaped control in terms of Peak-to-Peak amplitude of velocity error, which can achieve 72~\% while using input-shaping filters.}
\end{abstract}

\section{Introduction}

Cable-Driven Parallel Robots (CDPRs) are a particular class of parallel robots, where the rigid links are replaced by cables. They consist of a moving-platform connected to a base with cables. {Thanks to their large payload capacity, their high dynamic performance and large workspace with respect to their dimensions, CDPRs can be used for several types of industrial applications such as additive manufacturing \cite{izard2018improvements}, painting, sand blasting \cite{gagliardini2016discrete} and assembly \cite{pott2010large}. As safety fulfillments are taken into consideration for most of CDPR applications, these manipulators can also be used in search and rescue \cite{merlet2010portable} and rehabilitation \cite{homma2002study} operations.
}

CDPRs can reach high velocities and accelerations in large workspaces thanks to their low inertia \cite{lamaury2013dual}. However, vibrations may occur. Pose stabilization and/or trajectory tracking of the moving-platform can be degraded due to cable elasticity. Considering the physical cable characteristics, the cable elasticity has mainly two origins. The first one is the axial stiffness of the cables, which is associated with the elastic material modulus and the cable structure. The second is the sag-introduced flexibility, which comes from the effect of cable weight onto the static cable profile \cite{irvine1981max, yao2013modeling}.

CDPR accuracy improvement is still possible once the robot is manipulated through a suitable control scheme. Several controllers have been proposed in the literature to improve CDPR accuracy locally {or along a given trajectory} \cite{jamshidifar2015adaptive, fang2004motion, zi2008dynamic, de2018out}. 
{Some papers deal} with the CDPR control while considering cable elongations and their effect on the dynamic behavior. In~\cite{khalilpour2017wave}, an approach of wave based control (WBC) is proposed for large scale robots whose cables sagging effect cannot be neglected. It combines the position control and the active vibration damping simultaneously. This control strategy assumes actuator motion as launching a mechanical wave into the flexible system, which is absorbed on its return to the actuator. The assumption of modeling cables as elastic straight {and} massless links is valid for robots with relatively small size \cite{baklouti2017dynamic}. The cable mass {is supposed to be negligible} with respect to the moving-platform mass~\cite{diao2009vibration, kraus2013investigation}. The dynamic modeling and adaptive control of a single degree-of-freedom flexible cable-driven parallel robot (CDPR) is investigated in~\cite{ASME_DS_2019_GCF}. A Rayleigh-Ritz cable model is developed that takes into account the changes in cable mass and stiffness due to its winding and unwinding around the actuating winch, with the changes distributed throughout the cables. A robust H$_{\infty}$ control scheme for CDPR is described in \cite{laroche2013preliminary} while considering the cable elongations into the dynamic model of the moving-platform and cable tension limits. Besides, H$_{\infty}$ control scheme for position control of 6-DOF CDPRs is proposed in~\cite{chellal2014h}. Compared to~\cite{laroche2013preliminary}, {the position} control scheme is done in the operational space and the tension management is made separately in a more efficient way.
The control of CDPRs while assuming flexibility in cables is proposed in~\cite{IFAC2020_PTPCC, IROS2018_PCCP, ASME_DETC2018_PCPC}. In~\cite{khosravi2014dynamic, khosravi2016stability, khosravi2011dynamic}, the undesirable vibrations {are attenuated} by using the singular perturbation theory, which is based on measured cable elongations. Usually, cables are modeled by linear axial springs, with a constant stiffness. In~\cite{khosravi2015dynamic}, variable cable stiffness, which is a function of cable length variation is considered into the control loop. In this context, authors of \cite{babaghasabha2015fully} have also proposed a robust adaptive controller to attenuate vibrations in presence of kinematic and dynamic uncertainties. This control method requires {the measurement of cable lengths and moving-platform pose.}
However, the use of exteroceptive sensors to measure the moving-platform pose increases the complexity of the overall system~\cite{cui2013closed}.

The importance of the feed-forward effect on non-linear systems control is highlighted in~\cite{slotine1991applied}. It leads to stable systems with enhanced trajectory tracking performances. Feed-forward model-based controllers are used to fulfill accuracy improvement by using a CDPR reference model~\cite{zhang2017dynamic}. This latter predicts the mechanical behavior of the robot; and then generates an adequate reference signal to be followed by the CDPR. This type of control provides the compensation of the desirable effects without exteroceptive measurements. A model-based control scheme for CDPR used as a high rack storage is presented in~\cite{bruckmann2013design}. This research work takes into account the mechanical properties of cables, namely their elasticity. This strategy, integrating the mechanical behavior of cables in the reference signal, enhances the CDPR performance. Although cable stiffness is considered, a limitation of this control method lies in the cable interactions with the overall system that are not considered, which can result in unwanted oscillations of the moving-platform. A novel model-based control strategy allowing the pre-compensation not only of cable elongations but also the effect of their interaction with the overall system was proposed in \cite{baklouti2019elasto, baklouti2019vibration}. This control method reduces the oscillations of the moving-platform. 

To reduce the moving-platform vibrations, a frequency dependent method, named as input-shaping \cite{singhose2009command}, was proposed for the closed-loop control of flexible systems. Input-shaping method consists in re-designing the desired command signal {such that} the robot self-cancels residual vibrations. It is used for manipulators with flexible joints to remedy the resulting oscillations. {Input shaping was used for the control of serial robots} \cite{park2006design, aribowo2010input, zhao2016zero} such as the industrial SCARA manipulator \cite{ha2015wireless}. Oscillation control by shaping the input signal was also applied for conventional parallel robots \cite{kozak2004locally, li2009vibration, oltjen2016reduction}. Input-shaping was also used for the control of humanoid robots \cite{kobayashi2016investigation, yi2016vibration, rupert2015comparing} to deal with unwanted vibrations resulting from their non-rigid behavior. Once control methods are not allowing all DOF of the moving-platform of an under-actuated CDPR to be controlled, the extra DOFs in motion of the moving-platform makes it easy to sway \cite{yanai2001inverse}. This leads to low operational efficiency and then the lost of controllability. Input-shaping method for under-actuated CDPRs was proposed as an alternative to attenuate residual vibrations \cite{hwang2016trajectory, barry2016modeling, lin2016design, montgomery2017suppression, park2013anti}. Input-shaping was also proposed in~\cite{yoon2016multi} for over-actuated CDPRs. However, this control does not allow to manage the actuation redundancy. CDPR vibration reduction approach using input-shapers is presented in \cite{korayem2018tracking}. This control method integrates a feedback controller, whose linearization is based on CDPR rigid model.

Accordingly, this paper deals with the input-shaping for feed-forward control of CDPRs as a mean to attenuate the oscillations of their moving-platform. The main contribution of this paper lies in the use of input-shaping filters in conjunction with an experimentally validated model-based feed-forward control \cite{baklouti2019vibration} for disturbance rejection, while considering the overall stiffness of the system.
This control method integrates cable tension calculation to satisfy positive cable tensions along the prescribed trajectory of the moving-platform.

This paper is organized as follows: Section~\ref{sec:modeling} describes the CDPR parameterization and gives the equations required to establish the control law. Section~\ref{sec:IS_principle} introduces the principle and the design of input-shaping filters. Then, the application of input-shaping for feed-forward closed-loop control is presented in Section~\ref{sec:application_IS}. Experimental validations performed on the CREATOR prototype located at LS2N, Nantes, France, a CDPR with three cables and three Degree-Of-Freedom (DOF), are discussed in Section~\ref{sec:experimental}. The setup of the input-shapers is based on a robustness analysis as presented in Section~\ref{sec:Robustness_error}. {Finally, the effectiveness, in terms of moving-platform oscillation attenuation, of the proposed closed-loop control method combined with shaping inputs is experimentally studied in Section~\ref{Sec:results}.}

\section{CDPR parametrization and modeling } 
\label{sec:modeling}
\begin{figure}[!htb]
	\centering
	\includegraphics[width=0.5\linewidth]{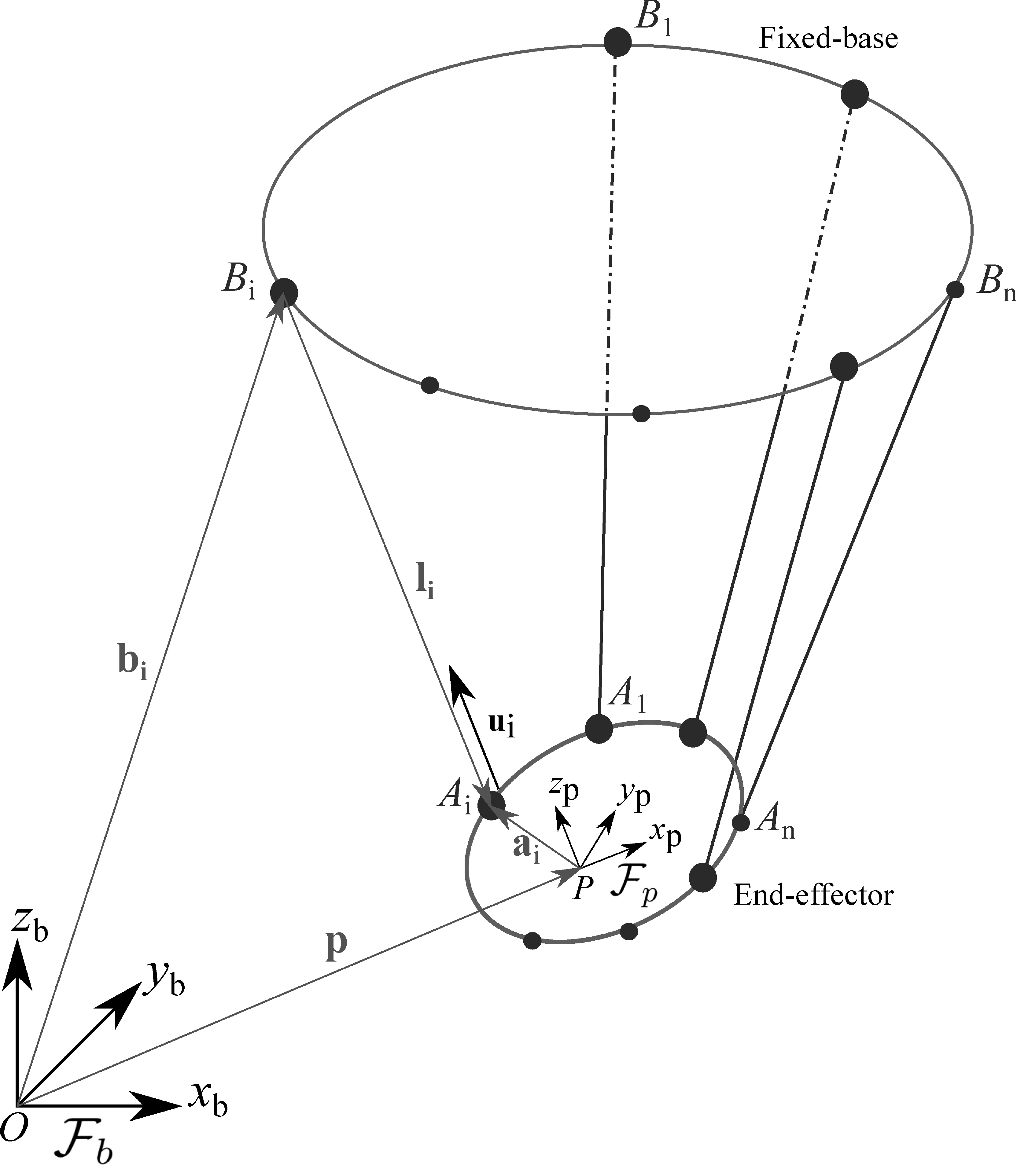}
	\caption{The $i$th closed-loop of a CDPR}
	\label{fig:reperes}
\end{figure}

A CDPR with $n$ cables, whose geometry is illustrated in Fig.~\ref{fig:reperes}, has $n$ anchor points, denoted by $A_i$ and $n$ exit points, denoted by $B_i$, $i\in[1~~n]$. The Cartesian coordinate vectors of anchor points $A_i$ and exit points $B_i$ of a CDPR, $i\in[1~~n]$, are denoted $\mathbf{a}_i$ and $\mathbf{b}_i$, respectively. These vectors are expressed in the moving-platform frame $\displaystyle{\mathcal{F}_p~=~\{P,~x_p,~y_p,~z_p\}}$ and in the base frame $\displaystyle{\mathcal{F}_b~=~\{O,~x_b,~y_b,~z_b\}}$, respectively.
 
The pose~$\displaystyle{\mathbf{x}~=~[\mathbf{p}^T~\mathbf{o}^T]^T\in \mathbb{R}^{m}}$ of the moving-platform geometric center $P$ in the base frame $\mathcal{F}_b$ is described by the position vector $\displaystyle{\mathbf{p}~=~[x,~y,~z]^T\in \mathbb{R}^{u}}$ and the orientation vector $\displaystyle{\mathbf{o}~=~[\phi,~\theta,~\psi]^T\in \mathbb{R}^{v}}$. The orientation of the moving-platform is parameterized by Euler angles~$\phi$, $\theta$ and $\psi$. $u$ being the number of translational Degrees-Of-Freedom (DOFs), $v$ being the number of rotational DOFs and $m$ being the total DOFs of the moving-platform.
The geometric closed-loop equations for a CDPR with straight line cables take the form:	
\begin{equation}
{\mathbf{l}_i}={\mathbf{b}_i}-{\mathbf{a}_i}-{\mathbf{p}},~~i=[1~n], \label{eq:geo}
\end{equation}
 where $\displaystyle{\mathbf{{l}}=[l_1,~...,~l_n]^T\in \mathbb{R}^{n}}$ is the cable length vector and $l_i=\|\mathbf{l}_{i}\|_2$ is the $i$th cable length. The inverse kinematics of the CDPR is expressed as follows: 
	\begin{equation}
\mathbf{\dot{l}}=\mathbf{A} {\mathbf{t}},~~	\mathbf{A}=\begin{bmatrix}
\mathbf{u}_1 & ~ &{\mathbf{a}_1}\times \mathbf{u}_1\\
.&~&.\\
.&~&.\\
\mathbf{u}_n  & ~&{\mathbf{a}_n}\times \mathbf{u}_n 
\end{bmatrix}, \label{eq:L_pt}
	\end{equation}
where $\mathbf{A}\in \mathbb{R}^{n\times m}$ is the Jacobian matrix, which is a function of $\mathbf{x}$ and maps the moving-platform velocities to the cable velocity vector. $\displaystyle{\mathbf{u}_i\in \mathbb{R}^{u}}$ is unit vector of $i$th cable pointing from $A_i$ to $B_i$. $\mathbf{t}=[\mathbf{\dot{p}}^T~\boldsymbol{\omega}^T]\in \mathbb{R}^{m}$ is the moving-platform twist and $\boldsymbol{\omega}=[\omega_x,~\omega_y,~\omega_z]^T$ is its angular velocity vector.

The actuator angular displacement and velocity vectors are denoted as $\displaystyle{\mathbf{q}~=~[q_1,~...,~q_n]^T\in \mathbb{R}^{n}}$ and \\
$\displaystyle{\dot{\mathbf{q}}~=~[\dot{q}_1,~...,~\dot{q}_n]^T\in \mathbb{R}^{n}}$, respectively. The relationship between $\mathbf{q}$ ($\mathbf{\dot{q}}$, resp.) and $\mathbf{l}$ ($\mathbf{\dot{l}}$, resp.) is linear:
	\begin{equation}
\mathbf{q}= \boldsymbol{\chi}^{-1}~\mathbf{{l}},~~~  \mathbf{\dot{q}}= \boldsymbol{\chi}^{-1}\mathbf{\dot{l}},
	\end{equation}
where $\displaystyle{\boldsymbol{\chi}~=~diag[\chi_1,~...,~\chi_n]\in \mathbb{R}^{n \times n}}$ is a diagonal matrix presenting the winches winding ratio.

The equations of motions are derived from the equations of Newton-Euler: 
\begin{equation}
\mathbf{M}~\mathbf{\dot{t}}+\mathbf{C}~\mathbf{t}
=\mathbf{W}\boldsymbol{\tau}+\mathbf{w}_{ex}, \label{eq:pfd}
\end{equation}
with 
\begin{equation}
\mathbf{M}=\begin{bmatrix}
m_{ee}\mathbb{I}_{m}& -m_{ee}\mathbf{d}^{\times}\\ 
m_{ee}\mathbf{d}^{\times}& \mathbf{I}_{ee}-m_{ee}\mathbf{d} ^{\times}\mathbf{d}^{\times}
\end{bmatrix}, \, \, 
\mathbf{C}~\mathbf{t}=\begin{bmatrix}
m_{ee}\boldsymbol{\omega}^{\times}\boldsymbol{\omega}^{\times}\mathbf{d}\\
\boldsymbol{\omega}^{\times}\left( \mathbf{I}_{ee}-m_{ee}\mathbf{d}^{\times}\mathbf{d}^{\times}\boldsymbol{\omega}\right) 
\end{bmatrix}, \, \,
\mathbf{W}=-\mathbf{A}^T.
\end{equation}
$\mathbf{\dot{t}}=[\mathbf{\ddot{p}}^T~\mathbf{\dot{r}}^T]\in \mathbb{R}^{m}$ is the acceleration vector of the moving-platform. $m_{ee}$ is the total mass of the moving-platform. $\mathbf{d}=\overrightarrow{PG}=[d_x,~d_y,~d_z]^T$ is the vector pointing from the moving-platform geometric center $P$ to mass center $G$.
$\mathbf{M}\in \mathbb{R}^{m\times m}$ is the generalized mass matrix of the moving platform\footnote{\begin{equation*}
	\boldsymbol{\omega}^{\times}=\begin{bmatrix}
	0&-\omega_z&\omega_y\\
	\omega_z&0&-\omega_x\\
	-\omega_y&\omega_x&0
	\end{bmatrix},~~\mathbf{{d}}^{\times}=\begin{bmatrix}
	0&-d_z&d_y\\
	d_z&0&-d_x\\
	-d_y&d_x&0
	\end{bmatrix}
	\end{equation*}.}. $\mathbf{I}_{ee}$ denotes the inertia matrix of the moving-platform expressed at its center of mass. $\mathbf{C}\in \mathbb{R}^{m\times m}$ is the matrix of Coriolis and centrifugal forces. $\mathbf{w}_{ex}\in \mathbb{R}^{m}$ is the external wrench applied onto the moving-platform. $\boldsymbol{\tau}=[\tau_1,~..,~\tau_n]^T\in \mathbb{R}^{n}$ is the cable tension vector, $\tau_i$  being the tension in cable $\mathcal{C}_i$, $i~=~[1..n]$. \\

The calculation of cable tension vector $\boldsymbol{\tau}$ presents two cases as a function of the degree of the CDPR actuation redundancy. 
The first case is when the number of cables $n$ is equal to the DOF $m$. In this case, the determination of cable tensions leads to only one solution as long as the wrench matrix $\mathbf{W}$ is not singular.
The second case arises when $m$~$<$~$n$. The redundancy allows to select a solution amongst the infinite set cable tension vectors as long as the moving-platform pose is wrench feasible\footnote{A cable force distribution is said to be feasible in a particular configuration and for a specified set of wrenches, if the tension forces in the cables can counteract any external wrench of the specified set applied to the moving-platform \cite{ebert2004connections}.}. 

\subsection{Stiffness modeling}

{For many CDPRs, the axial stiffness of the cable, which is a function of its elastic modulus, is the main source of flexibility, which must be considered into the CDPR model \cite{baklouti2018sensitivity}. Admitting cable tension model (either linear, \textit{i.e.} directly proportional to cable elongation \cite{baklouti2019elasto}, or non-linear, \textit{i.e.} a logarithmic function of cable elongation \cite{baklouti2017dynamic}), the stiffness matrix is expressed as a function of the moving-platform pose, the CDPR geometry and the cable tensions.}	

Assuming all cables are sufficiently tensioned, and submitted to low strains and small elongations, we can consider the linear cable tension model: 
 \begin{equation}
\boldsymbol{\tau}=\mathbf{K}_l\delta\mathbf{l}. \label{Eq:tension_lin}
\end{equation}
$\displaystyle{\mathbf{K}_l=diag[k_1,~...,~k_n]\in \mathbb{R}^{n \times n}}$ being a diagonal matrix presenting the $i$th cable axial stiffness $k_i=\dfrac{EA}{l_i}$.\\ $\displaystyle{\delta\mathbf{l}=[\delta l_1,~...,~\delta l_n]^T\in \mathbb{R}^{n}}$ is the cable elongation vector. $E$ is the cable modulus of elasticity and $A$ is the cross-section of the cable. 

{As $\delta\mathbf{l} =\mathbf{A}\delta \mathbf{x}$, the generalized overall stiffness matrix $ \mathbf{K}_x$ of the CDPR is expressed as follows: 
\begin{equation}
\mathbf{K}_x=\mathbf{A}^T\mathbf{K}_l\mathbf{A}.
\end{equation}
$\delta \mathbf{x}$ being the moving-platform displacement screw corresponding to the cable elongation vector $\delta\mathbf{l}$ from their static equilibrium configuration. Therefore, the moving-platform free vibration around any equilibrium is managed by the following linear perturbation equation:
\begin{equation}
\mathbf{M}~\delta \mathbf{\dot{t}}+{\mathbf{C}}~\delta \mathbf{t} + \mathbf{K}_x \, \delta \mathbf{x} 
=\mathbf{0}. \label{eq:pfd1}
\end{equation}}

\subsection{Natural frequencies}

As the input-shaping method is frequency dependent, the knowledge of natural frequencies of the CDPR is essential. The natural frequencies of the CDPR can be calculated by solving the classic eigenvalue problem based on the global stiffness matrix of the CDPR as mentioned in \cite{kozak2006static}. {They are obtained by solving the generalized eigenvalue problem associated with the generalized mass matrix $\mathbf{M}$ and stiffness matrix $\mathbf{K}_x$, which is described as follows:}
\begin{equation}
\mathrm{det}({\lambda} \mathbb{I}_{m} - \mathbf{M}^{-1}\mathbf{K}_x)=0, \label{eq:frequence_prop}
\end{equation}
{where $\mathbb{I}_{m}\in \mathbb{R}^{m \times m}$ is the identity matrix. The $i$th natural frequency $f_i$~(Hz) corresponds to the $i$th solution ${\lambda}_i$ of the characteristic polynomial described by Eq.~(\ref{eq:frequence_prop}), $f_i=\dfrac{\sqrt{{\lambda}_i}}{2\pi}$.}

Note that the natural frequencies can be calculated considering the cable damping through the use of a complex Young modulus. In that case, Dynamic Mechanical analysis (DMA) can be used to identify the cable stiffness and cable damping under forced oscillatory measurements. In \cite{baklouti2017dynamic}, it was shown that the elastic modulus and damping thus obtained are highly dependent on frequency for a given preload, over a representative frequency range of the CDPR behavior.

\section{Input-shaping for feed-forward control}
\label{sec:feedforward}
This section deals with the principle and the design of input-shaping filters leading to the reduction of residual vibrations by generating a self-canceling control signal. The application of these filters into the closed-loop feed-forward control of CDPRs is then introduced.  

\subsection{Input-shaping Principle}
\label{sec:IS_principle}
Input shaping principle consists of a convolution of the control inputs with series of impulses, each described by an impulse amplitude and a delay \cite{singh2002input}. Those coefficients are chosen in a way that the sum of residual vibrations produced by each impulse cancel each other and produces a reference trajectory slightly different from the original one, which does not produce residual vibrations of the system. 

\begin{figure}[!htb]
	\centering
	\includegraphics[width=1\linewidth]{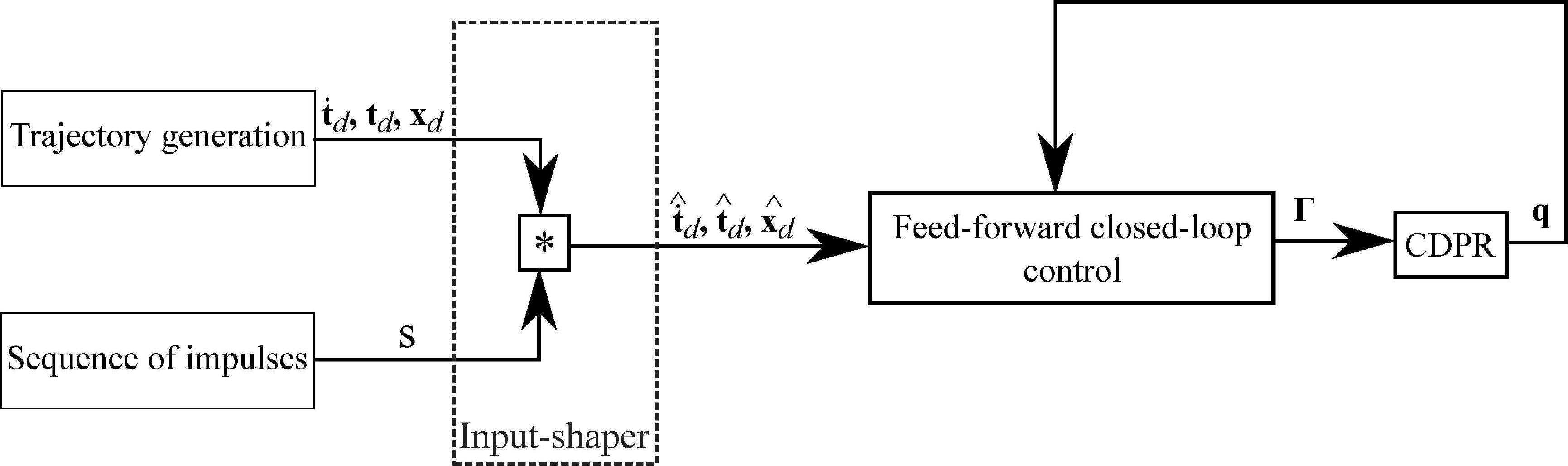}
	\caption{Block diagram: Input-shaping for closed-loop control scheme}
	\label{fig:controlinputshapinggeneral1}
\end{figure}

Figure~\ref{fig:controlinputshapinggeneral1} shows the considered control architecture. The input shaper filters modify the desired moving-platform trajectory (pose $\mathbf{x}_d$, velocity $\mathbf{t}_d$ and acceleration $\mathbf{\dot{t}}_d$) and provides a shaped trajectory (pose $\hat{\mathbf{x}}_d$, velocity $\hat{\mathbf{t}}_d$ and acceleration $\hat{\mathbf{\dot{t}}}_d$), which becomes the desired motion of the moving-platform. If the input shaper is properly designed, then the new shaped command will lead to a response for which the residual vibration is reduced or eliminated as explained in \cite{barry2016modeling}.

Two types of input-shapers are distinguished: single-mode and multi-mode input-shapers. {The behavior of a $k$-mode input shaper can be described by the applied sequence of impulses $\mathbf{S}$ defined by the following transfer function:}
\begin{equation}
\mathbf{S}(p)=\dfrac{\hat{\mathbf{x}}_d (p)}{\mathbf{x}_d(p)}=\dfrac{\hat{\mathbf{t}}_d (p)}{\mathbf{t}_d(p)}=\dfrac{\hat{\dot{\mathbf{t}}}_d (p)}{\dot{\mathbf{t}}_d(p)}=\sum_{i=1}^{k}\mathbf{D}_i e^{-pt_i},
\end{equation}
where $k$ is the number of considered vibration modes into input-shaping filters design. $p$ is the Laplace variable. $\mathbf{D}_i$ is the $i$th mode impulse amplitude vector and is a function of the natural frequency $f_i$. Its corresponding damping ratio is $\zeta_i$. $t_i$ is the time location of the applied impulse. \\

Zero-Vibration (ZV) shaper is a single mode input-shaper. It amounts to the convolution of the original input signal with a sequence of two impulses. The impulses are separated by the half of the robot's natural period. The ZV-shaper for $i$th natural mode is expressed as follows:
\begin{equation}
\mathbf{ZV}_i=\begin{bmatrix}
\mathbf{D}_i \\
\mathbf{t}_i
\end{bmatrix}
=
\begin{bmatrix}
\dfrac{1}{1+\epsilon_i}& \dfrac{\epsilon_i}{1+\epsilon_i} \\
0&\dfrac{1}{2f_i}
\end{bmatrix},~~\epsilon_i=e^{ \dfrac{-\zeta_i \pi}{\sqrt{1-\zeta_i^2}}}.
\label{Eq:k_is}
\end{equation}

Zero-Vibration-Derivative (ZVD) is a single mode input-shaper. It consists in the convolution of the original input signal with a sequence of three impulses separated by half the period of the robot's vibration. The ZVD-shaper for $i$th natural mode is expressed as follows:
\begin{equation}
\mathbf{ZVD}_i=\begin{bmatrix}
\mathbf{D}_i \\
\mathbf{t}_i
\end{bmatrix}
=
\begin{bmatrix}
\dfrac{1}{1+2\epsilon_i+\epsilon_i^2}& \dfrac{2\epsilon_i}{1+2\epsilon_i+\epsilon_i^2}& \dfrac{\epsilon_i^2}{1+2\epsilon_i+\epsilon_i^2} \\
0&\dfrac{1}{2f_i}&\dfrac{1}{f_i}
\end{bmatrix}.
\end{equation}

A convolution of multiple single mode shapers aims to control multiple modes of a robot manipulator \cite{tokhi2008flexible}. A simple way to obtain a two-mode shaper is to convolve two single-mode shapers \cite{singhose1997convolved}. In fact, the ZV-ZV shaper (ZVD-ZVD shaper, resp.) consists of the convolution of two ZV shapers (ZVD shapers, resp.) applied for $i$th and $j$th natural modes: 
\begin{equation}
\mathbf{ZV-ZV}_{i-j}=\begin{bmatrix}
\mathbf{D}_{i-j} \\
\mathbf{t}_{i-j}
\end{bmatrix}=	
\begin{bmatrix}
\dfrac{ 1}{(\epsilon_i + 1)(\epsilon_j + 1)}&
\dfrac{\epsilon_j}{(\epsilon_i + 1)(\epsilon_j + 1)}&
\dfrac{ \epsilon_i}{(\epsilon_i + 1)(\epsilon_j + 1)}&
\dfrac{\epsilon_i\epsilon_j}{(\epsilon_i + 1)(\epsilon_j + 1)}
\\
0& \dfrac{1}{2f_j}& \dfrac{1}{2f_i}& \dfrac{1}{2(f_i+f_j)}
\end{bmatrix}
\end{equation}
and
\begin{equation}
\begin{split} 
\mathbf{ZVD-ZVD}_{i-j}=\begin{bmatrix}
\mathbf{D}_{i-j} \\
\mathbf{t}_{i-j}
\end{bmatrix},\\
\mathbf{D}_{i-j} =\dfrac{1}{(\epsilon_i + 1)^2(\epsilon_j + 1)^2} \begin{bmatrix}
{1}~~~&
{2\epsilon_j}~~~&
{\epsilon_j^2}~~~&
{2\epsilon_i}~~~&
{4\epsilon_i\epsilon_j}~~~&
{2\epsilon_i\epsilon_j^2}~~~&
{\epsilon_i^2}~~~&
{2\epsilon_i^2\epsilon_j}~~~&
{\epsilon_i^2\epsilon_j^2}
\end{bmatrix},\\
\mathbf{t}_{i-j}=\begin{bmatrix}0~~~& \dfrac{1}{2f_j}~~~&  \dfrac{1}{f_j}~~~& \dfrac{1}{2f_i}~~~&\dfrac{f_i+f_j}{2f_if_j}~~~& \dfrac{2f_i+fj}{2f_if_j}~~~& \dfrac{1}{f_i}~~~& \dfrac{2f_j+f_i}{2f_jf_i}~~~& \dfrac{f_i+f_j}{f_if_j}
\end{bmatrix}
\end{split}
\end{equation}

The integration of an input-shaping filter leads to oscillation attenuation, but not to cancellation even if a multi-mode shaper is used. This fact can be related to uncertainties in the CDPR model leading to frequency identification errors.

\subsection{Application to feed-forward control}
\label{sec:application_IS}
\begin{figure}[!htb]
	\centering
	\includegraphics[width=1\linewidth]{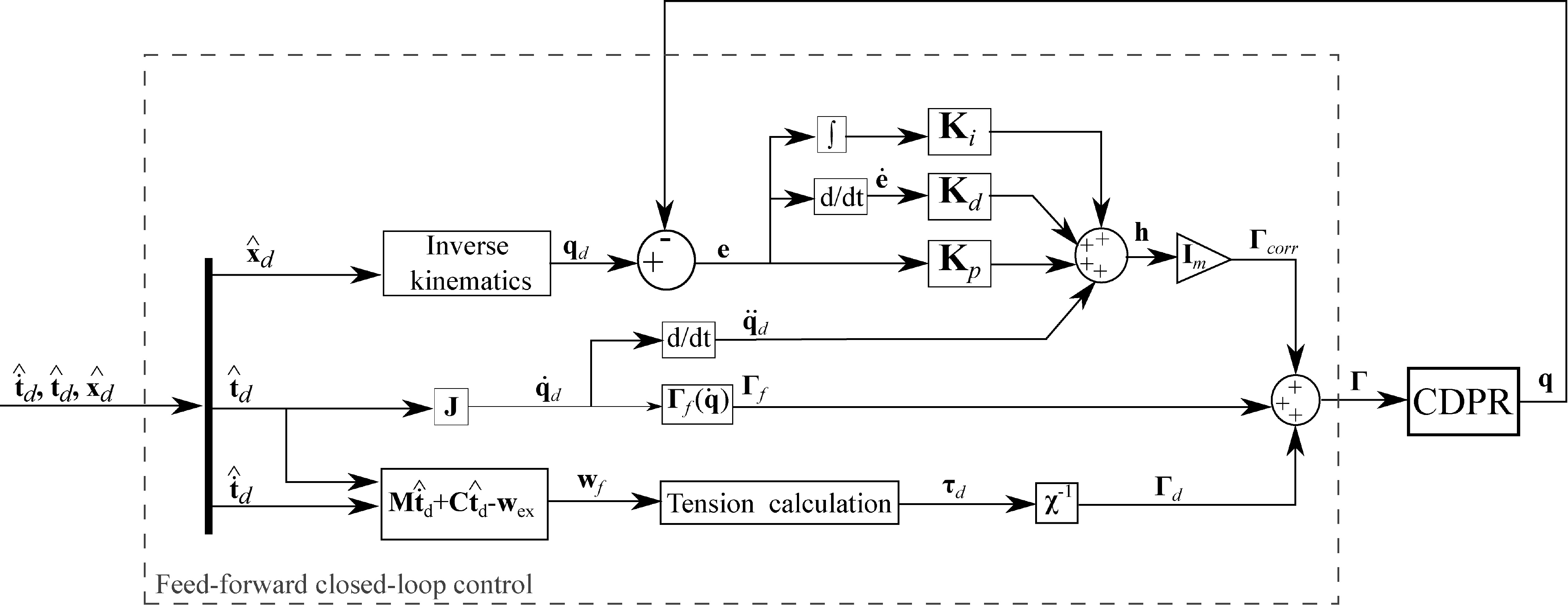}
	\caption{Feed-forward closed-loop control scheme}
	\label{fig:controlinputshapinggeneral}
\end{figure}

Figure~\ref{fig:controlinputshapinggeneral} presents the application of input-shaping into closed-loop feed-forward control, {based on a classical PID control}. The shaped motion of the moving-platform (pose $\hat{\mathbf{x}}_d$, velocity $\hat{\mathbf{t}}_d$ and acceleration $\hat{\dot{\mathbf{t}}}_d$) becomes the desired one for the moving-platform. The vectors $\mathbf{q}_{d}$, $\mathbf{\dot{q}}_{d}$ and $\mathbf{\ddot{q}}_{d}$ present the winch angular displacement vector, the winch velocity vector and the winch acceleration vector, respectively, corresponding to the shaped motion of the moving-platform (pose $\hat{\mathbf{x}}_d$, velocity $\hat{\mathbf{{t}}}_d$ and acceleration $\hat{\dot{\mathbf{t}}}_d$). 

The torque set-point $\boldsymbol{\Gamma}$ assigned to the motors is expressed as:
\begin{equation}
\boldsymbol{\Gamma}=\boldsymbol{\Gamma}_{corr}+\boldsymbol{\Gamma}_f(\mathbf{\dot{q}}_{d})+\boldsymbol{\Gamma}_{d},
\label{Eq:winch_closed}
\end{equation}
where $\boldsymbol{\Gamma}_{d} =\boldsymbol{\chi}^{-1} \boldsymbol{\tau}_{d}  \in \mathbb{R}^n$ is the torque set-point corresponding to the tension calculated along the shaped trajectory. $\mathbf{w}_{f}\in \mathbb{R}^{m}$ denotes the external wrench and is the input signal of the tension calculation block. It is calculated based on the generated trajectory and the external wrench applied on the moving-platform, 
$\displaystyle{\mathbf{w}_f= \mathbf{M}\hat{\dot{\mathbf{t}}}_d+\mathbf{C}\hat{\mathbf{{t}}}_d-\mathbf{w}_{ex}.}$
It is about solving the dynamic equilibrium equations for a given pose~$\mathbf{x}$ of the moving-platform, which can be described as follows:
\begin{equation}
\mathbf{W}\boldsymbol{\tau}_d+\mathbf{w}_{f}=\mathbf{0},
\label{eq:eq_statique}
\end{equation} 
\\
The cables can only pull and not push the moving-platform. If the number of cables $n$ is equal to the DOF $m$, the inversion of Eq.~(\ref{eq:eq_statique}) is possible, as long as the wrench matrix $\mathbf{W}$ is not singular.
However, this tension set can present negative values. That means that one or more cable(s) may have to push the moving-platform, which is not possible. When~$m < n$, Eq.~(\ref{eq:eq_statique}) may have an infinite number of solutions. Therefore, the redundancy allows to select a solution amongst the infinite set cable tension vectors satisfying some criteria. The problem of force distribution presents one important control issue for redundant actuated CDPRs, which is the determination of feasible cable force distribution\footnote{A cable force distribution will be said to be feasible in a particular configuration and for a specified set of wrenches if the cable tensions can counteract any external wrench of the specified set applied to the moving-platform \cite{ebert2004connections}.} \cite{gosselin2011determination, mikelsons2008real,rasheed:hal-02379201,CableCon2017RLMC}.

$\boldsymbol{\Gamma}_f(\mathbf{\dot{q}}_{d})$ is the joint friction torque vector, which is expressed according to the static model of friction \cite{khalil2004modeling} as follows: 
\begin{equation}
\boldsymbol{\Gamma}_f(\mathbf{\dot{q}}_{d})= \boldsymbol{\Gamma}_s~ \textrm{sgn}(\mathbf{\dot{q}}_{d})+\boldsymbol{\Gamma}_v~\mathbf{\dot{q}}_{d},
\label{Eq:static_friction}
\end{equation}
where 
$\boldsymbol{\Gamma}_s\in \mathbb{R}^{n \times n}$ is a diagonal matrix containing the dry friction coefficients and $\boldsymbol{\Gamma}_v\in \mathbb{R}^{n \times n}$ is a diagonal matrix containing the viscous friction coefficients.\\

$\displaystyle{\boldsymbol{\Gamma}_{corr}~=~\mathbf{I}_m \mathbf{h}(t)}$ corresponds to the torque of correction, where $\mathbf{I}_m$ is a diagonal matrix containing the winch moment of inertia. $\mathbf{h}(t)$ is defined by:
\begin{equation}
\mathbf{h}(t)=\mathbf{\ddot{q}}_{d}+\mathbf{K}_{p}~(\mathbf{q}_{d}-\mathbf{q}) +\mathbf{K}_{d}~(\mathbf{\dot{q}}_{d}-\mathbf{\dot{q}})+\mathbf{K}_{i}~\int_{t}^{t^+} (\mathbf{q}_{d}-\mathbf{q})~dt,
\end{equation}
where $\mathbf{K}_{p}\in \mathbb{R}^{n\times n}$ is the proportional gain matrix, $\mathbf{K}_{d}\in \mathbb{R}^{n\times n}$ is the derivative gain matrix, $\mathbf{K}_{i}\in \mathbb{R}^{n\times n}$ is the integrator gain matrix. $\mathbf{{q}}$ is the measured angular displacement vector of motors. The closed-loop dynamics corresponds to the following tracking error equation:
\begin{equation}
\mathbf{\ddot{e}}+\mathbf{K}_{p}~\mathbf{e} +\mathbf{K}_{d}~\mathbf{\dot{e}}+\mathbf{K}_{i}~\int_{t}^{t^+} \mathbf{e}~dt= \mathbf{0},
\label{eq:error_control}
\end{equation}
where $\mathbf{e}=\mathbf{q}_{d}-\mathbf{q}$ and $\mathbf{\dot{e}}=\mathbf{\dot{q}}_d-\mathbf{\dot{q}}$ are the tracking errors.

\section{Experimental setup and results}
\label{sec:experimental}
\begin{figure}[!htb]
	\centering
	\includegraphics[width=0.7\linewidth]{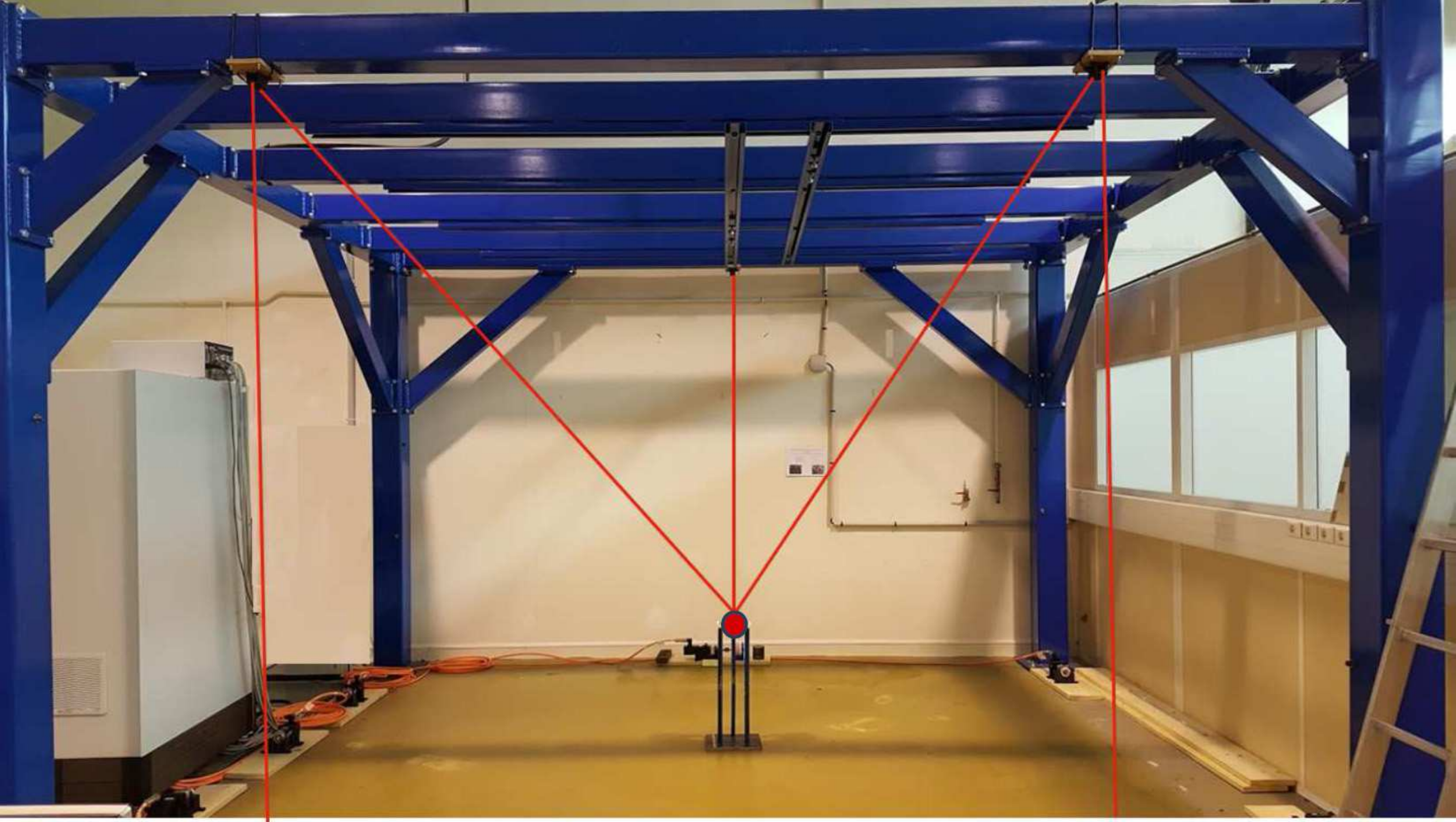}
	\caption{CREATOR prototype: LS2N, Nantes, France}
	\label{fig:cdprcreator}
\end{figure}

Experimentations have been performed here to analyze the effect of the integration of input-shaping filters into the closed-loop control scheme. A suspended configuration of the reconfigurable CREATOR prototype, shown in Fig.~\ref{fig:cdprcreator}, with 3 cables and 3 DOF is studied. The size of this prototype is about 4.5~m long, 4~m wide and 3~m high. The Cartesian coordinates of exit points $B_i$ expressed in~$\displaystyle{\mathcal{F}_b}$ are: $\mathbf{b}_1=[-2.085,~0.651,~2.726]^T m$, $\mathbf{b}_2=[ 2.085,~0.651,~2.735]^T m$ and $\mathbf{b}_3=[ -1.079,~ -1.898,~ 2.733]^T m$. 
Those coordinates were measured with a RADIAN Laser tracker\footnote{The static measurement accuracy of the used laser tracker is $\pm$~10~$\mu$m. Its range of working is between 0 and 25~m.}. The nominal mass of the moving-platform is equal to~0.650~kg. It is supposed to be a point-mass. The reader is referred to \cite{baklouti2019vibration} for more details about the identification procedure.

The CREATOR cables are made up of eight threads of polyethylene fiber with a diameter of 0.5~mm. These cables were experimentally upstream identified. The identification method is described in~\cite{baklouti2019vibration}. The absolute uncertainties in the applied force and resulting elongation measurements from the test bench outputs are estimated to be $\pm$~1~N and $\pm$~0.03~mm, respectively \cite{baklouti2017dynamic}. The resulting modulus of elasticity of the cable is equal to 70~$\pm$~1.51~GPa.

The CREATOR prototype is actuated by three Parker$^{\mbox{\scriptsize{\texttrademark}}}$ motors with gearboxes, whose moment of inertia 
$I_m$ is equal to 0.0031~kg.m$^2$. This set is connected to 3D printed winches. Each motor is connected to a Parker$^{\mbox{\scriptsize{\texttrademark}}}$ motor drive, which communicates with the dSpace$^{\mbox{\scriptsize{\texttrademark}}}$ controller through bi-directional real-time links. 

The friction torques of the actuators are identified with respect to the static friction model \cite{khalil2004modeling} by incrementing the joint angular velocity. As we do not have an accurate measurement of the motor torques, we suppose that the viscous friction torque is null. Here, the dry friction value is equal to 0.14~N.m.

\begin{figure}[!htb]
	\centering
	\includegraphics[width=0.7\linewidth]{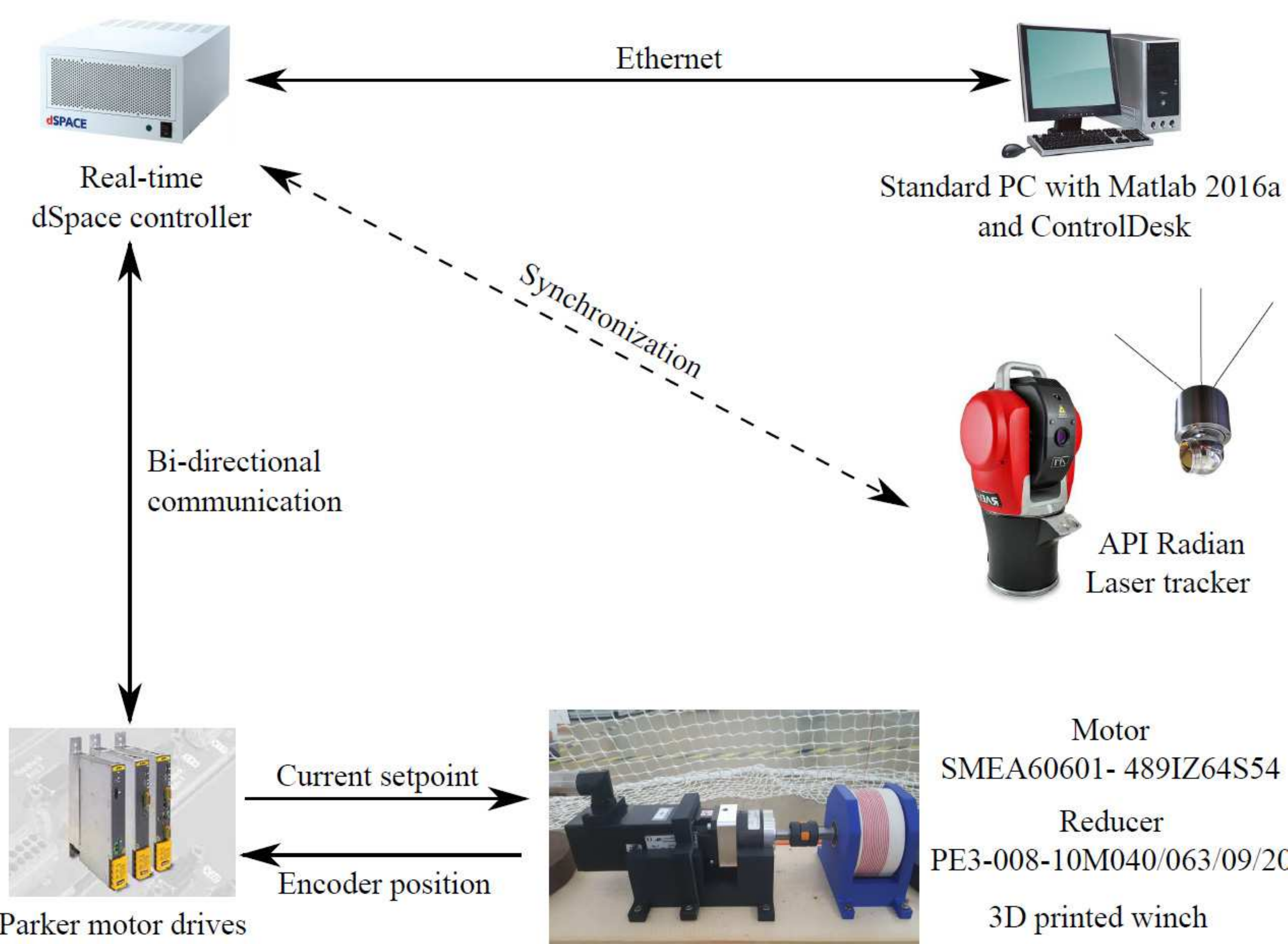}
	\caption{Main hardware components of the CREATOR protype}
	\label{fig:equivalentcreator}
\end{figure}

The equivalent architecture of the CREATOR prototype is described in Fig.~\ref{fig:equivalentcreator}.
The command of the CREATOR prototype is implemented in a host PC through a software interface generated by ControlDesk$^{\mbox{\scriptsize{\textregistered}}}$\footnote{ControlDesk is the dSPACE experiment software for seamless electronic control unit development. It performs all the necessary tasks and gives a single working environment, from the start of experimentation right to the end.}. This latter enables the real-time simulation of the control schemes, created with Matlab-Simulink$^{\mbox{\scriptsize{\textregistered}}}$, in the dSpace$^{\mbox{\scriptsize{\texttrademark}}}$ control unit.

The used gains ${K}_{p}$, ${K}_{d}$ and ${K}_{i}$ of the PID controller are equal to 1125.8~N.m, 58.12~N.m and 7269.60~N.m, respectively. 

\subsection{Trajectory generation}
\label{sec:trj}
\begin{figure}[!htb]
	\centering
	\subfloat[]{\includegraphics[width=0.4\linewidth]{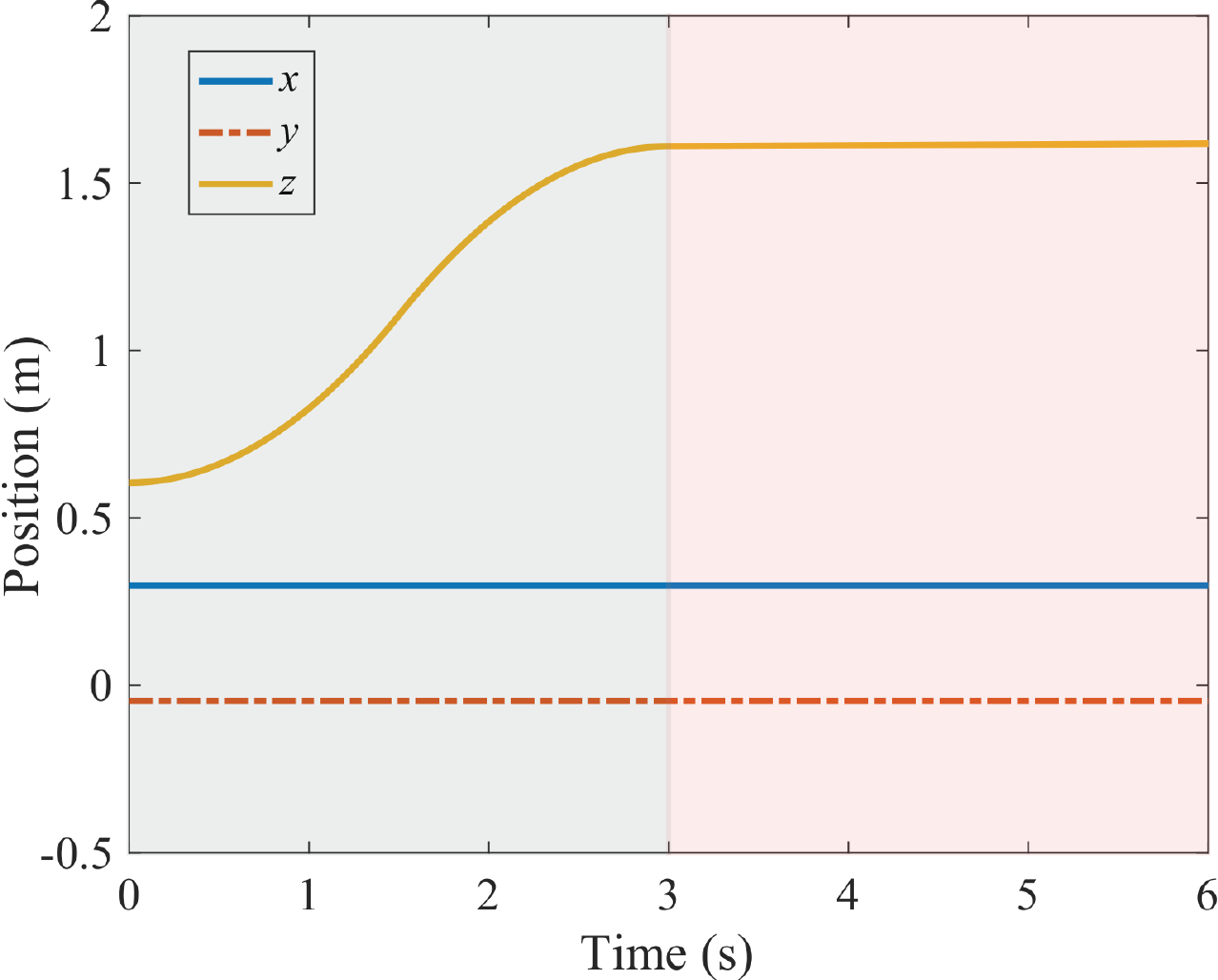} 	\label{fig:p_is}}
	\subfloat[]{\includegraphics[width=0.4\linewidth]{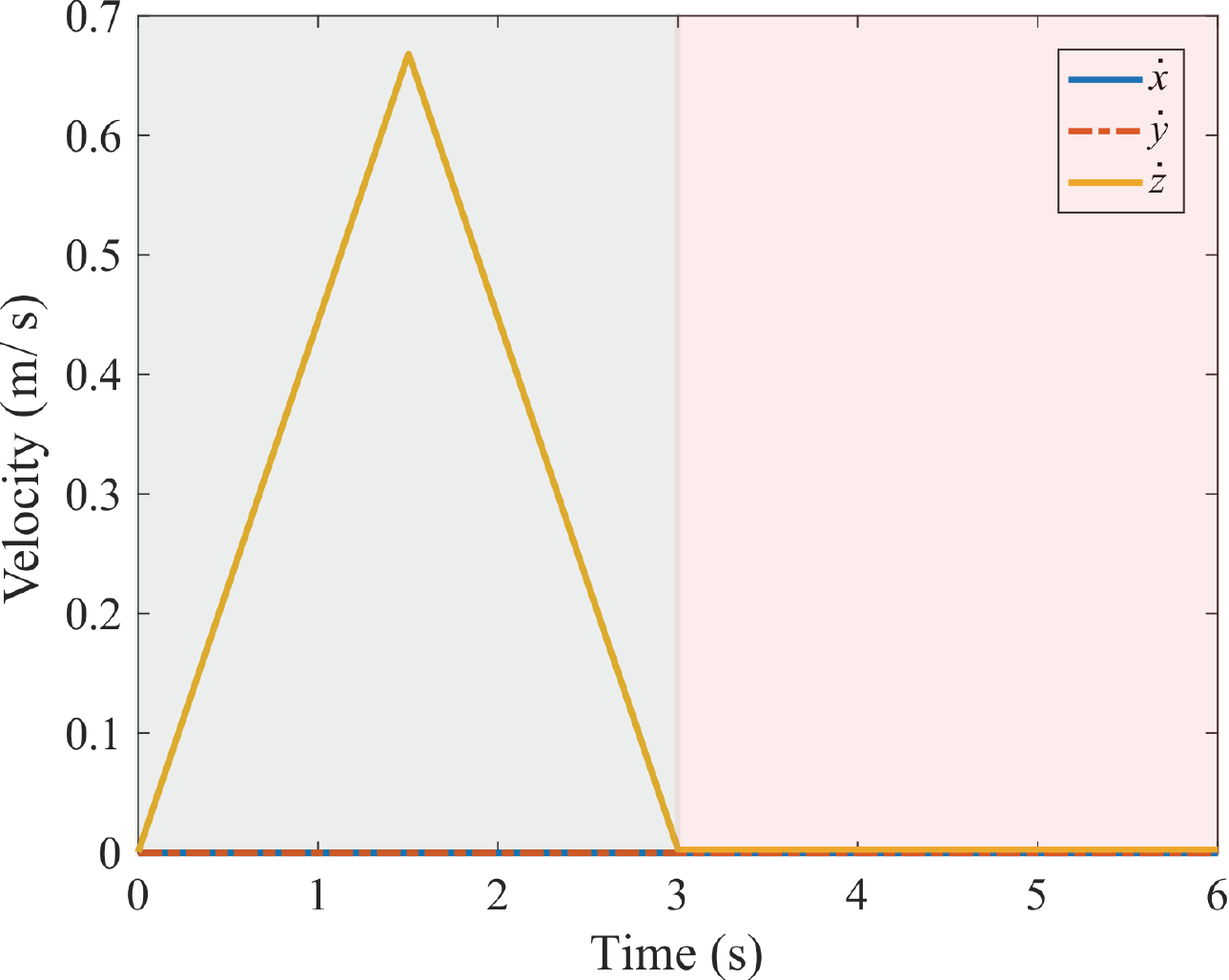}	\label{fig:dp_is}}\\
	\subfloat[]{\includegraphics[width=0.4\linewidth]{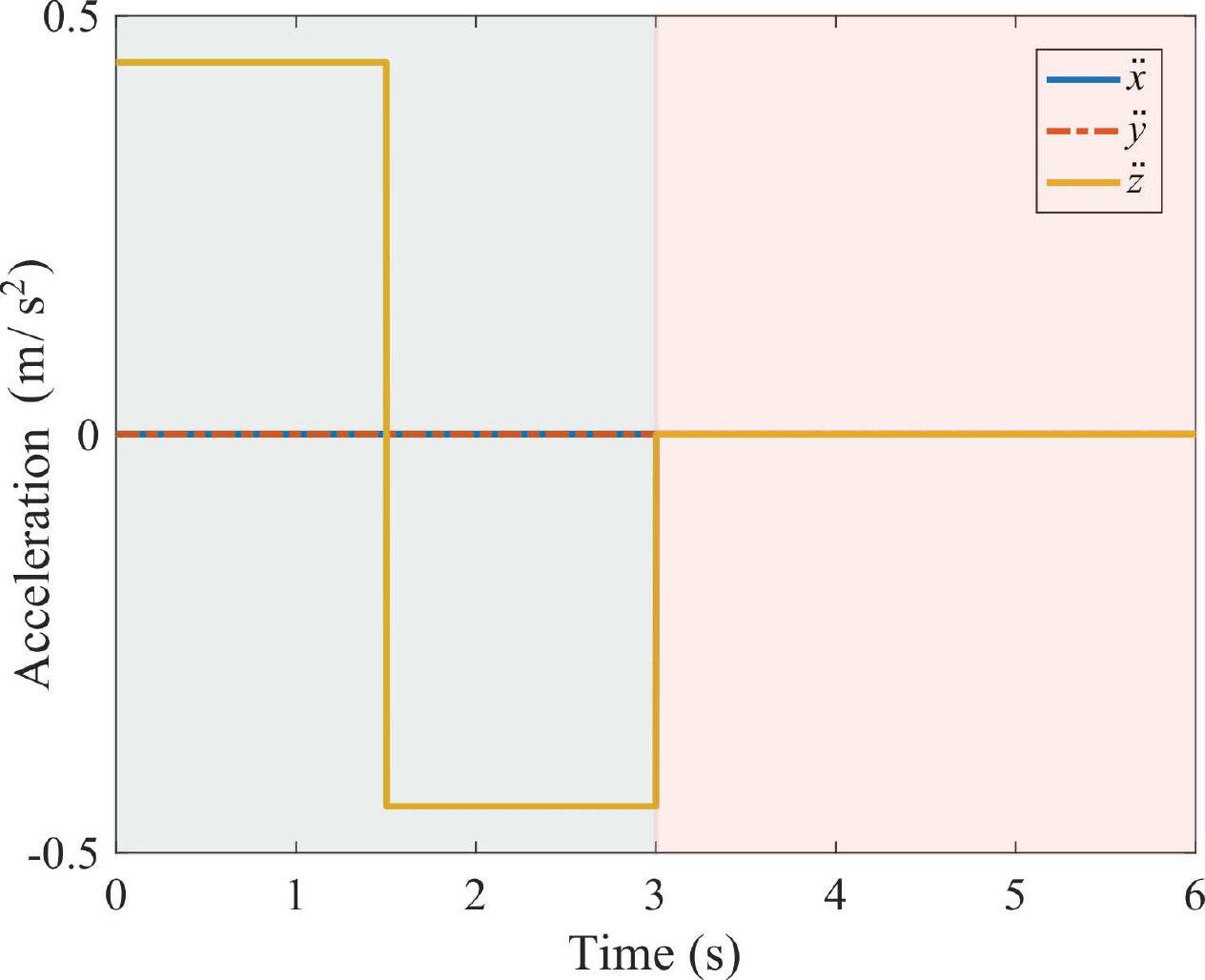}	\label{fig:ddp_is}}
	\caption{Nominal (a) position (b) velocity (c) acceleration profiles of the moving-platform}
\end{figure}

A non-smooth velocity trajectory is chosen to excite the natural modes of the CDPR. The motion is uniformly accelerated until the moving-platform achieves a desired position along $x$-axis, $y$-axis and $z$-axis. The accelerations of the moving-platform along $x$-axis, $y$-axis and $z$-axis are defined by a bang-bang profile (Fig.~\ref{fig:ddp_is}).

 The nominal trajectory of the moving-platform is a vertical straight line (Fig~\ref{fig:p_is}) from point $P_1$ of Cartesian coordinate vector $\displaystyle{{\bf p}_1~=~[0.29,-0.047,0.62]^T}$~m to point $P_2$ of Cartesian coordinate vector $\displaystyle{{\bf p}_2~=~[0.29,-0.047,1.62]^T}$~m during $t_f$~=~3~s. The nominal velocity (Fig.~\ref{fig:dp_is}) and acceleration (Fig.~\ref{fig:ddp_is}) profiles of the moving-platform are the time derivatives of the nominal trajectory. Therefore, the trajectory followed by the moving-platform is parametrized as follows: 
\begin{subequations}
	\begin{equation}
	\mathbf{p}(t)=\boldsymbol{\beta}_1 t^2 +\boldsymbol{\beta}_2t+\boldsymbol{\beta}_3,~~t\in[0~~\dfrac{t_f}{2}], 
	\end{equation}
	\begin{equation}
	\mathbf{p}(t)=\boldsymbol{\beta}_4(t-t_f)^2+\boldsymbol{\beta}_5(t-t_f)+\boldsymbol{\beta}_6,~~t\in[\dfrac{t_f}{2}~~{t_f}],
	\end{equation}
	\begin{equation}
	\mathbf{p}(t)=\mathbf{p}_2,~~t>t_f,
	\end{equation}
\end{subequations}
where 
\begin{equation}
\boldsymbol{\beta}_1=\boldsymbol{\beta}_4=\dfrac{2\left( \mathbf{p}_1-\mathbf{p}_2\right) }{t_f^2}, ~\boldsymbol{\beta}_3=\mathbf{p}_1, ~\boldsymbol{\beta}_6=\mathbf{p}_2, ~\boldsymbol{\beta}_2=\boldsymbol{\beta}_5=\mathbf{0}.
\end{equation}

\subsection{Input-shaper Robustness to Variations in Natural Frequency}
\label{sec:Robustness_error} 
The first calculated natural frequencies of the CREATOR prototype under study are $f_1$~=~3.67~Hz, $f_2$~=~6.34~Hz and $f_3$~=~7.82~Hz, which correspond to impulse times 0.27~s, 0.15~s and 0.12~s. Those natural frequencies are calculated at the initial pose of the moving-platform trajectory by solving the generalized eigenvalue problem associated with the apparent stiffness of the CDPR as mentioned in Section~\ref{sec:modeling}. Note that the natural frequencies are calculated while neglecting cable damping.
The residual vibrations are cancelled when the shaper parameters are perfectly tuned; conversely an error on the identification of the mode frequencies induces residual vibrations. Therefore, the shapers setup is preceded by a robustness analysis to errors in frequency identification.

\begin{figure}[!htb]
	\centering
	\includegraphics[width=0.5\linewidth]{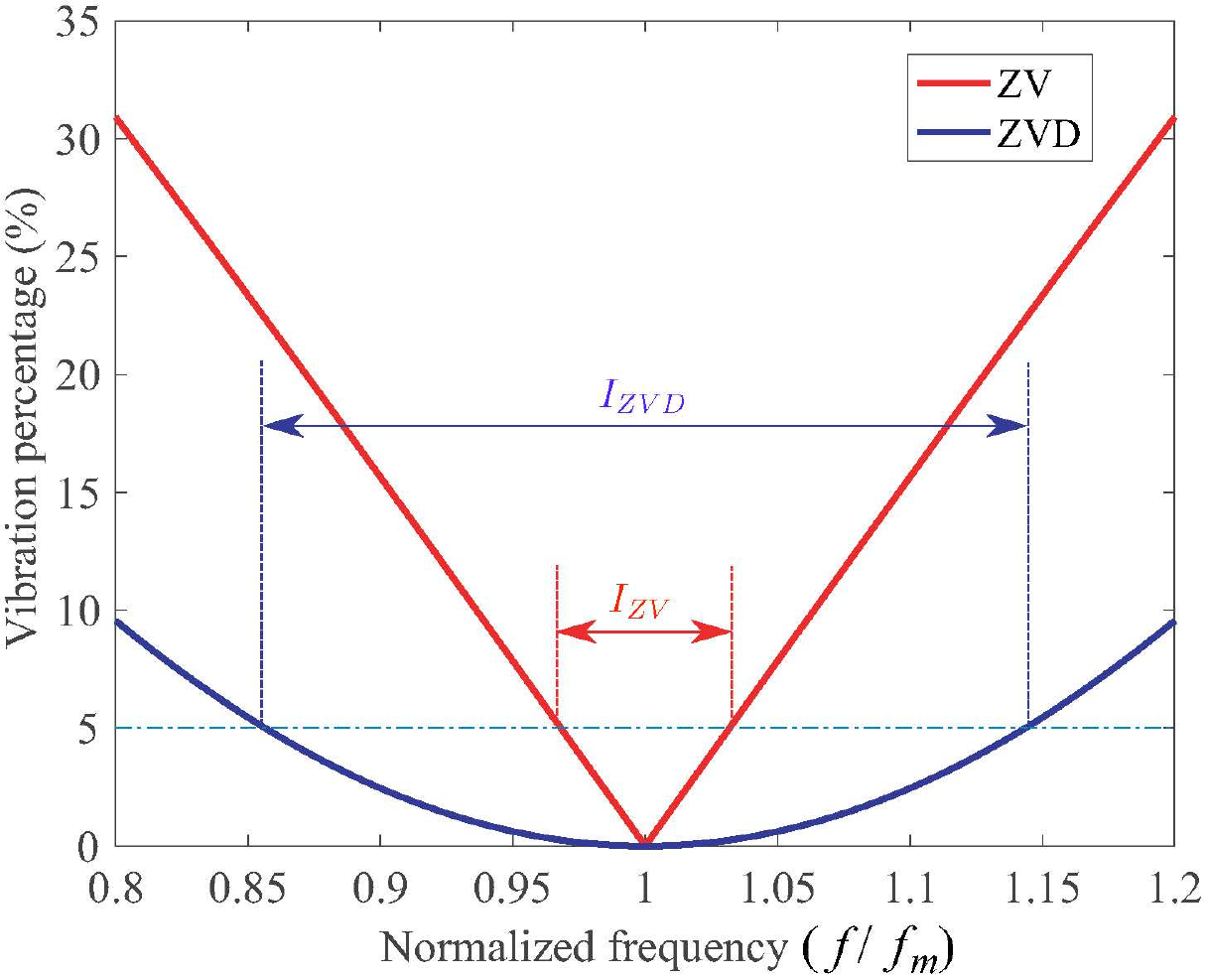}
	\caption{Sensitivity curves of ZV and ZVD input-shapers}
	\label{fig:sensitivityis}
\end{figure}

Figure~\ref{fig:sensitivityis} displays the sensitivity curves for single mode ZV and ZVD shapers~\cite{singer1990preshaping}. It shows the amplitude of residual vibrations as a function of the normalized frequency~$f/{f_m}$. $f_m$~is the predicted value of the natural frequency obtained from the model and $f$ is the effective natural frequency of the CDPR for a given end-effector pose. Here, the frequency~$f_m$ used to tune the input-shaper is set to $3.67$~Hz and the corresponding damping ratio $\zeta_1$ to zero. The  vibration percentage is the ratio in percentage between the amplitude of vibrations when input shaping is used and the amplitude of residual vibrations when shaping is not used. 

The sensitivity curve of the ZV-shaper shows that the larger the modeling errors, the faster the residual vibration increases when using a ZV-shaper. Besides, the vibrations obtained when using a ZVD shaper remain small. The robustness can be measured quantitatively by measuring the width of the curve at some low level of vibration percentage. This non-dimensional robustness measure is called the shaper insensitivity. The shaper insensitivity is denoted as~$I_{ZV}$ for the ZV-shaper and as~$I_{ZVD}$ for the ZVD-shaper.

The 5~\% shaper insensitivity is shown in Fig.~\ref{fig:sensitivityis} and is defined as a safety limit. For the ZV-shaper, $I_{ZV} = 0.06$ for the residual vibrations to remain smaller than 5~\% of the unshaped vibration. Similary, $I_{ZVD} = 0.28$. From Fig.~\ref{fig:sensitivityis}, it is noteworthy that the ZVD-shaper is significantly more robust than the ZV-shaper to modeling errors and to variations in the first natural frequency along the path followed by the robot end-effector.

The effective natural frequency of the CDPR can be measured by using a modal test with an impact hammer or dynamic shaker around an equilibrium position or by Fourier transform of the CDPR response during an exciting trajectory tracking. From~\cite{baklouti2019vibration}, the experimental value of the first natural frequency of CREATOR is equal to 3.6~Hz at the starting point of the test path, which is a value very close to the 3.67~Hz predicted by the model.

\begin{figure}[!htbp]
	\centering
	\includegraphics[width=0.7\linewidth]{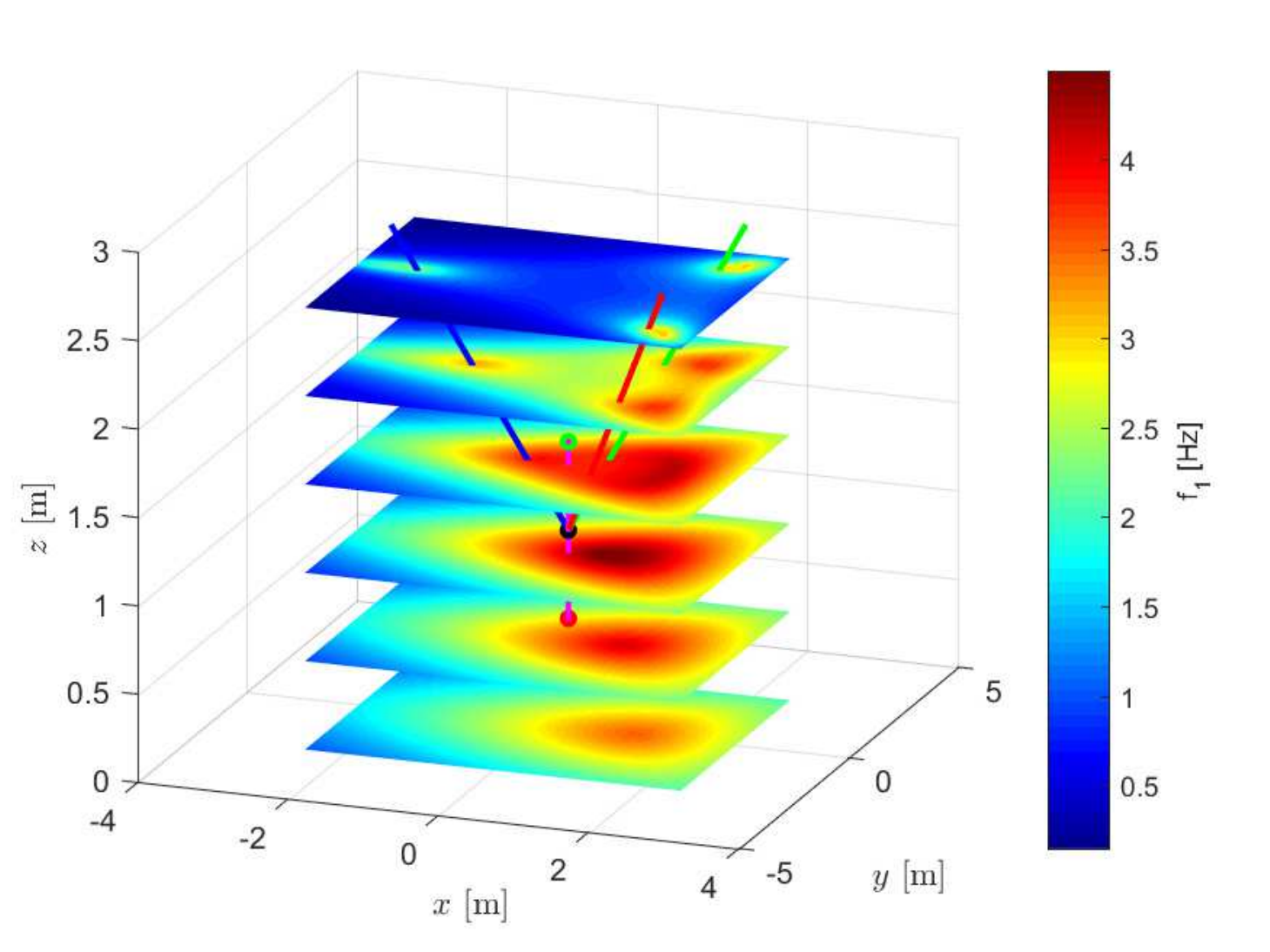}
	\caption{Calculated values of the first natural frequency of CREATOR through the Cartesian space and the test path (in magenta) followed by the end-effector}
	\label{fig:isworkspace}
\end{figure}

Figure~\ref{fig:isworkspace} shows the calculated values of the first natural frequency~$f_1$ of CREATOR prototype through the Cartesian space and the test path (in magenta) followed by its end-effector. A schematic of the CDPR with its effector in the middle of the test path is also depicted in Fig.~\ref{fig:isworkspace}. For better clarity, the contours of~$f_1$ are presented along slices spaced 0.5~m  apart along the $z$-axis.

\begin{figure}[!htbp]
	\centering
	\includegraphics[width=0.7\linewidth]{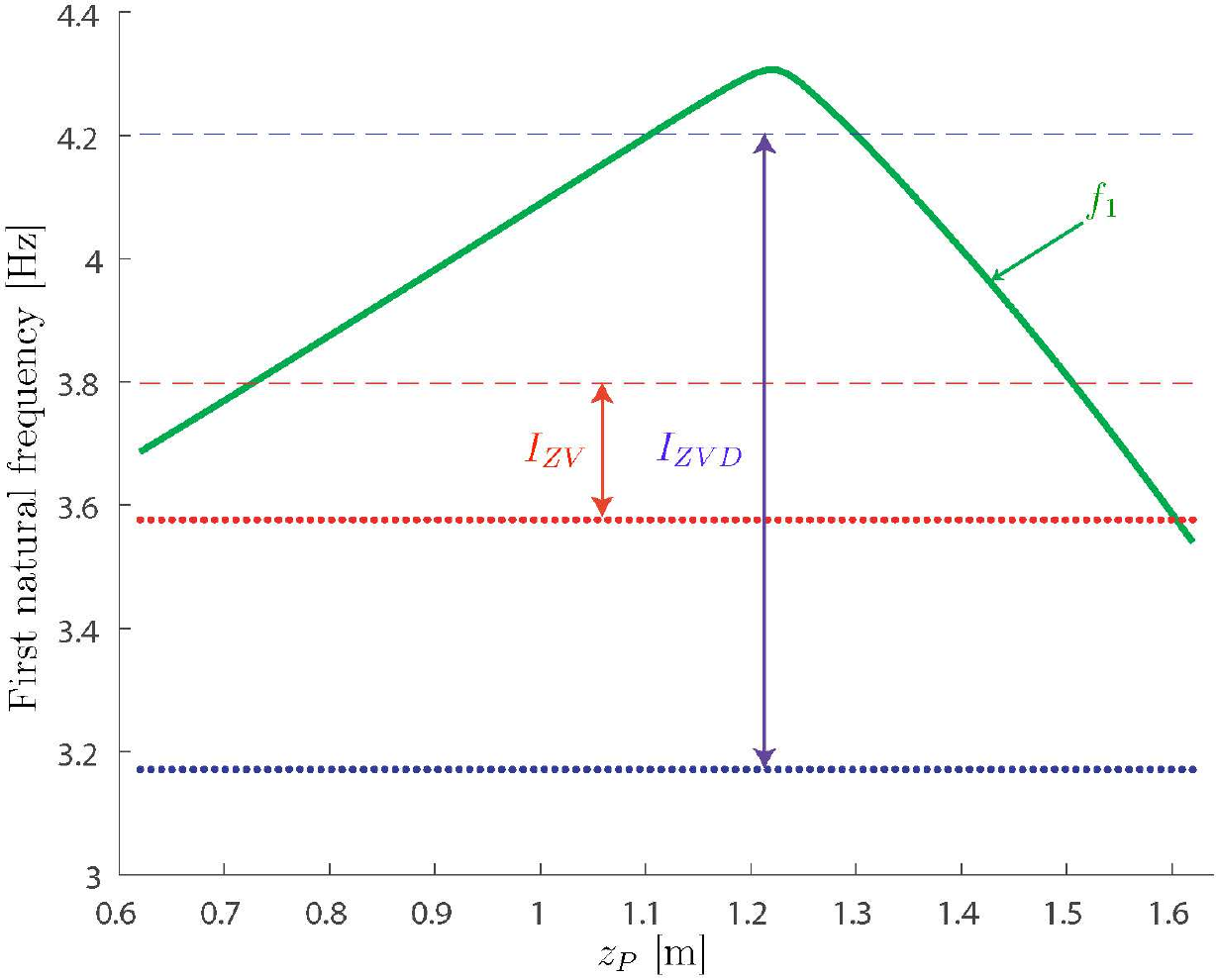}
	\caption{Calculated values of the first three natural frequencies of CREATOR along the test path and the bounds on the first natural frequency for the vibration percentage to remain smaller than 5\%}
	\label{fig:f1f2f3vsPath}
\end{figure}

Figure~\ref{fig:f1f2f3vsPath} shows the calculated values of~$f_1$ as a function of the $z$-coordinate, $z_P$, of the point-mass end-effector~$P$ along the test path. The shaper insensitivities~$I_{ZV}$ and~$I_{ZVD}$ are represented in Fig.~\ref{fig:f1f2f3vsPath} too. Input-shapers are considered robust to variations on the first natural frequency when the normalized frequency does not exceed the 5~\% insensitivity. Here, the normalized frequency $f_1/f_{m1}$ should remain between 0.97 (0.86, resp.) and 1.03 (1.14, resp.) while using a ZV shaper (ZVD shaper, resp.). Considering the value of the first natural frequency set to $f_{m1}$ = 3.67~Hz, the variation range to guarantee the 5~\% insensitivity is 0.22~Hz (from 3.57~Hz to 3.79~Hz) while using a single mode ZV-shaper. This same variation range increases to 1.03~Hz (from 3.17~Hz to 4.21~Hz) with a single mode ZVD-shaper. As one can see in Fig.~\ref{fig:f1f2f3vsPath}, the 5~\% insensitivity is practically respected over the sub-workspace for the ZVD-shaper, while the criterion is respected only for the beginning and the end of the trajectory for the ZV-shaper.

\begin{figure}[!htbp]
	\centering
	\subfloat[]{\includegraphics[width=0.5\linewidth]{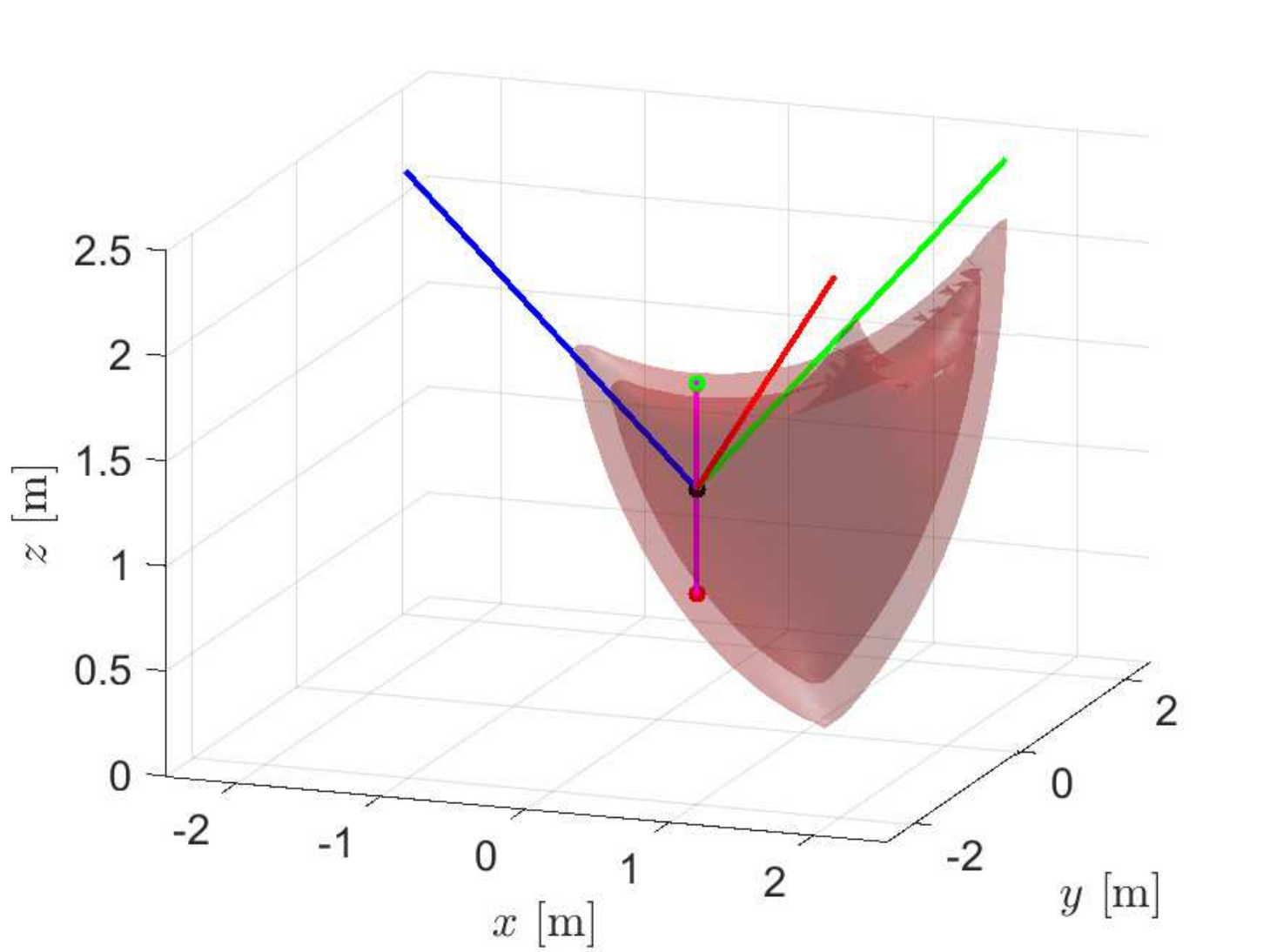}\label{fig:WFZV}}
	\subfloat[]{\includegraphics[width=0.5\linewidth]{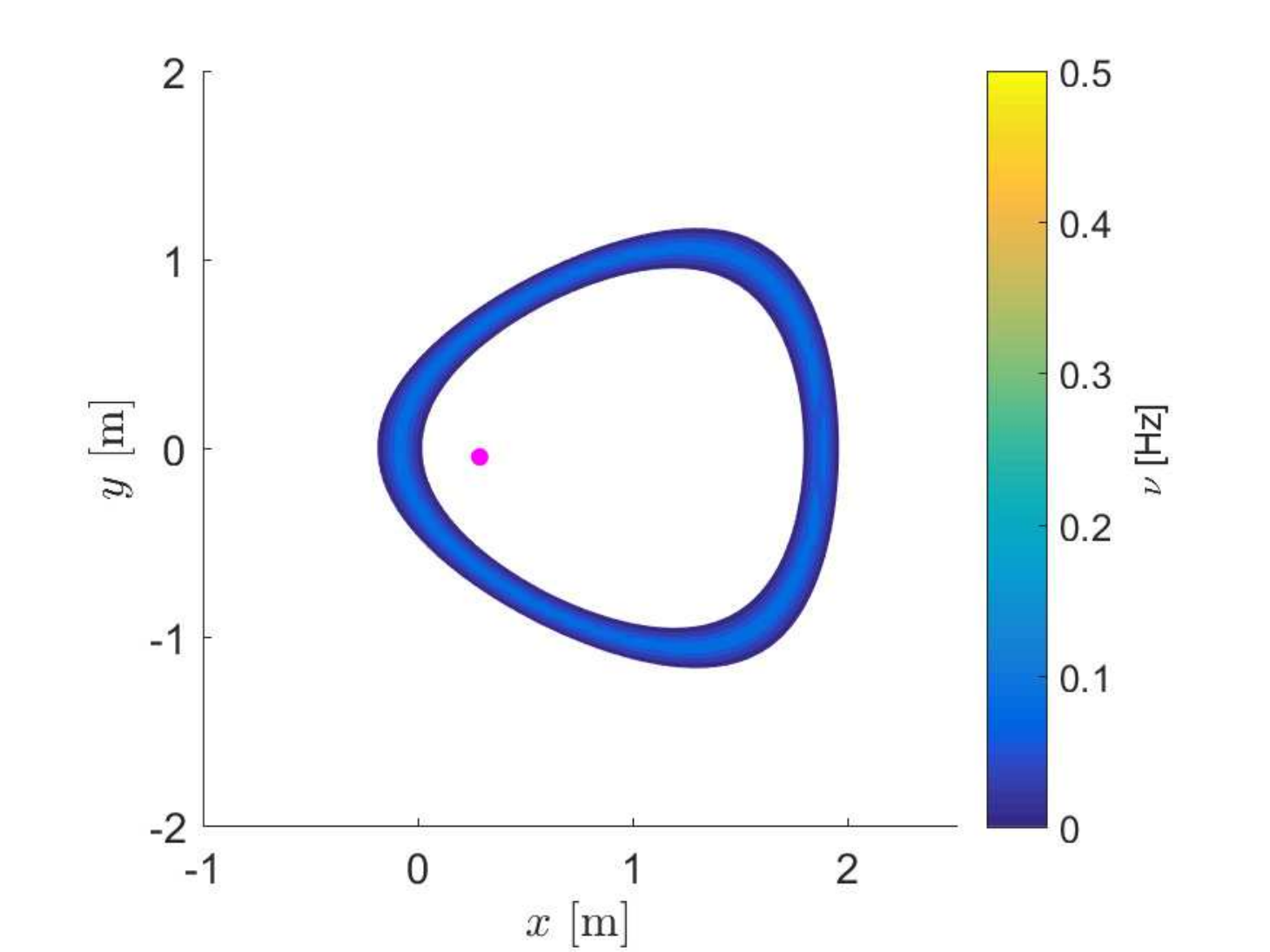}\label{fig:ContZV2D}}
    \label{fig:WFZV_Cont2D}
	\caption{The ZV-shaper sensivity workspace~$\mathcal{W}_{ZV}$ for CREATOR and the test path shown in magenta: (a)~3D view; (b)~Sectional view for $z_P = 1$~m}
\end{figure}

\begin{figure}[!htb]
	\centering
	\subfloat[]{\includegraphics[width=0.5\linewidth]{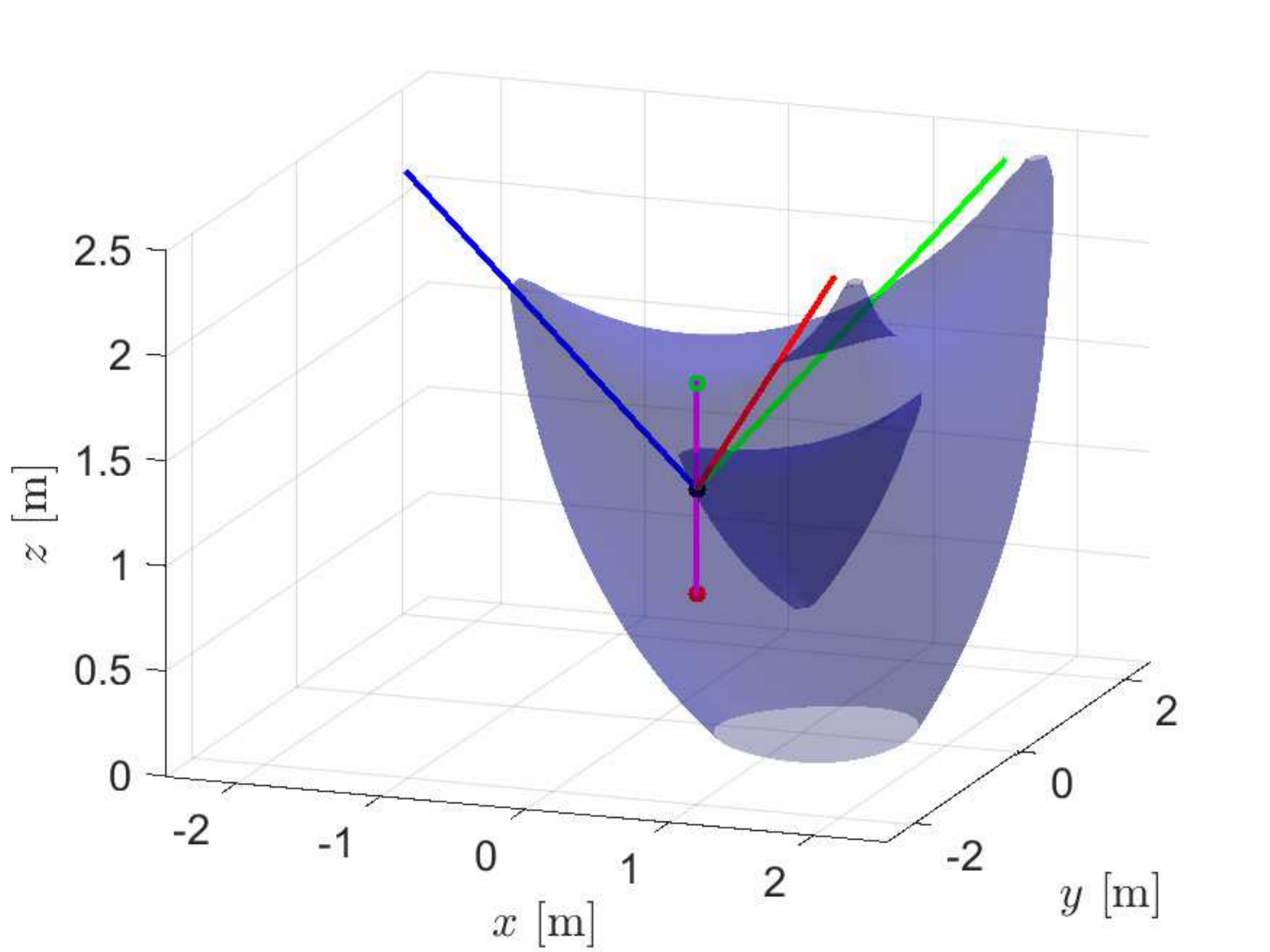}\label{fig:WFZVD}}
	\subfloat[]{\includegraphics[width=0.5\linewidth]{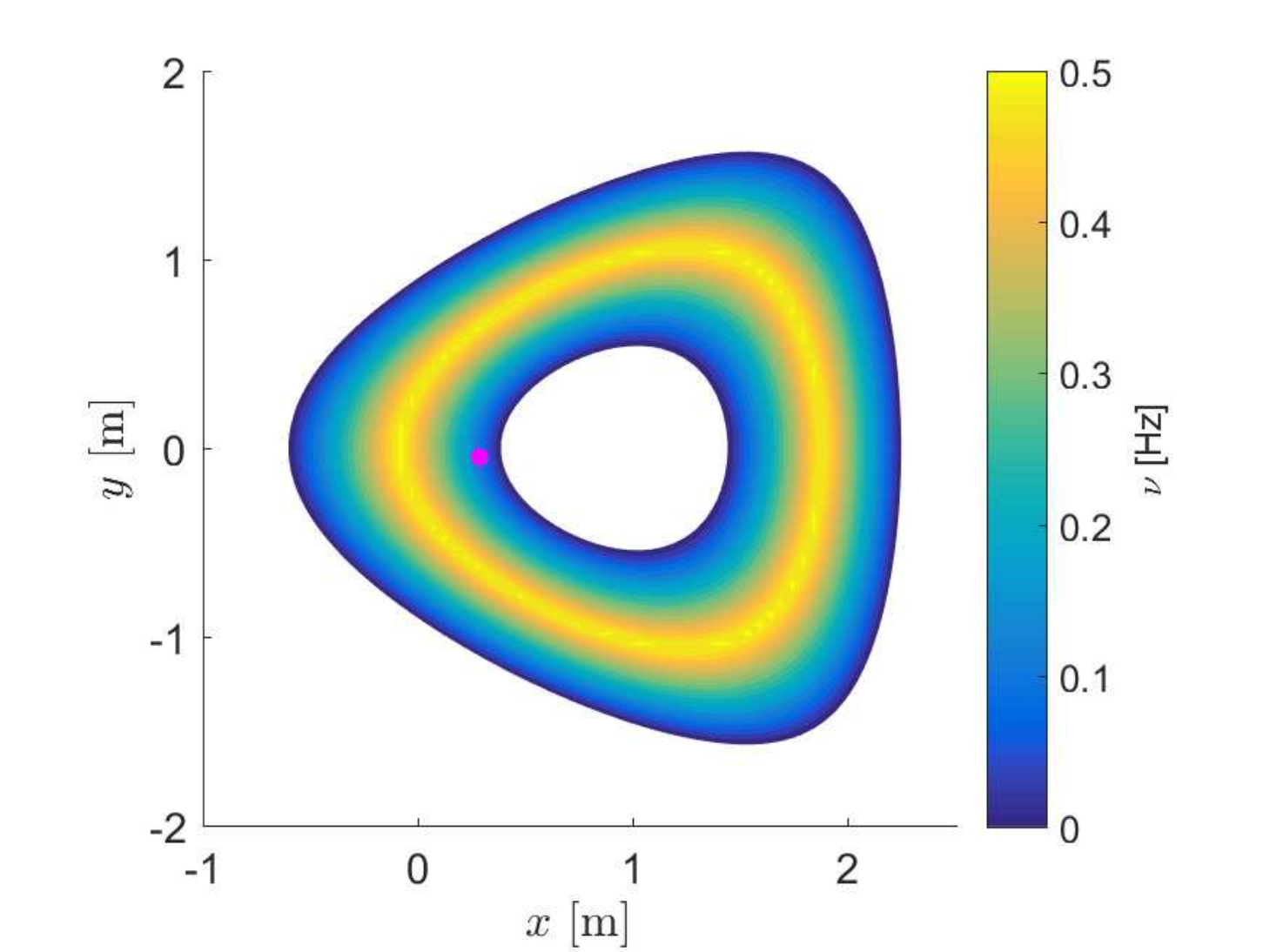}\label{fig:ContZVD2D}}
	\caption{The ZVD-shaper sensivity workspace~$\mathcal{W}_{ZVD}$ for CREATOR and the test path shown in magenta: (a)~3D view; (b)~Sectional view for $z_P = 1$~m}
\end{figure}

The ZV-shaper sensivity workspace~$\mathcal{W}_{ZV}$ associated to the ZV-shaper and the calculated value $f_{m1}$ of the first natural frequency at the starting point of the test path is shown in Fig.~\ref{fig:WFZV} and defined as:
 \begin{equation} \label{eq:WFZV}
\mathcal{W}_{ZV} = \{ {\bf p} \in \Rbb^{3}:  1 - \frac{I_{ZV}}{2} \leq \frac{f_1}{f_{m1}} \leq 1 + \frac{I_{ZV}}{2} \}
\end{equation}
with ${\bf p}$ denoting the Cartesian coordinate vector of the point-mass end-effector.
 
The ZVD-shaper sensivity workspace~$\mathcal{W}_{ZVD}$ associated to the ZVD-shaper and the calculated value $f_{m1}$ of the first natural frequency at the starting point of the test path is shown in Fig.~\ref{fig:WFZVD} and defined as

\begin{equation} \label{eq:WFZVD}
\mathcal{W}_{ZVD} = \{ {\bf p} \in \Rbb^{3}:  1 - \frac{I_{ZVD}}{2} \leq \frac{f_1}{f_{m1}} \leq 1 + \frac{I_{ZVD}}{2} \}
\end{equation}

The index~$\nu$ characterizing the robustness of the shaper to variations in~$f_1$ at a specific end-effector pose is defined as follows:
\begin{equation} \label{eq:nu}
\nu = \min \{f_1 - (1 - \frac{I_j}{2}) \, f_{m1} , (1 + \frac{I_j}{2}) \, f_{m1} - f_1 \} \, , \, j = \{ZV, \, ZVD \}
\end{equation}

Figure~\ref{fig:ContZV2D} (\ref{fig:ContZVD2D}, resp.) represents the contours of~$\nu$ through the sectional view of~$\mathcal{W}_{ZV}$ ($\mathcal{W}_{ZVD}$, resp.) at the $z$-coordinate $z_P = 1$~m. $\nu = 0$~Hz means that the end-effector is located on the boundary of the shaper sensivity workspace. The larger~$\nu$, the lower the vibration percentage of the end-effector~(Fig~\ref{fig:sensitivityis}). Figures.~\ref{fig:WFZV}-b and~\ref{fig:WFZVD}-b clearly show that the area where the path can be defined from the considered starting point is much larger with the ZVD-shaper than with ZV-shaper. It should be noted that an adaptive shaper~\cite{tzes1993adaptive, khorrami1995experimental, kojima2007adaptive} should be used once the path goes beyong the shaper sensitivity workspace.

\subsection{Experimental results}
\label{Sec:results}
\begin{figure}[!htb]
	\centering
	\subfloat[]{\includegraphics[width=0.45\linewidth]{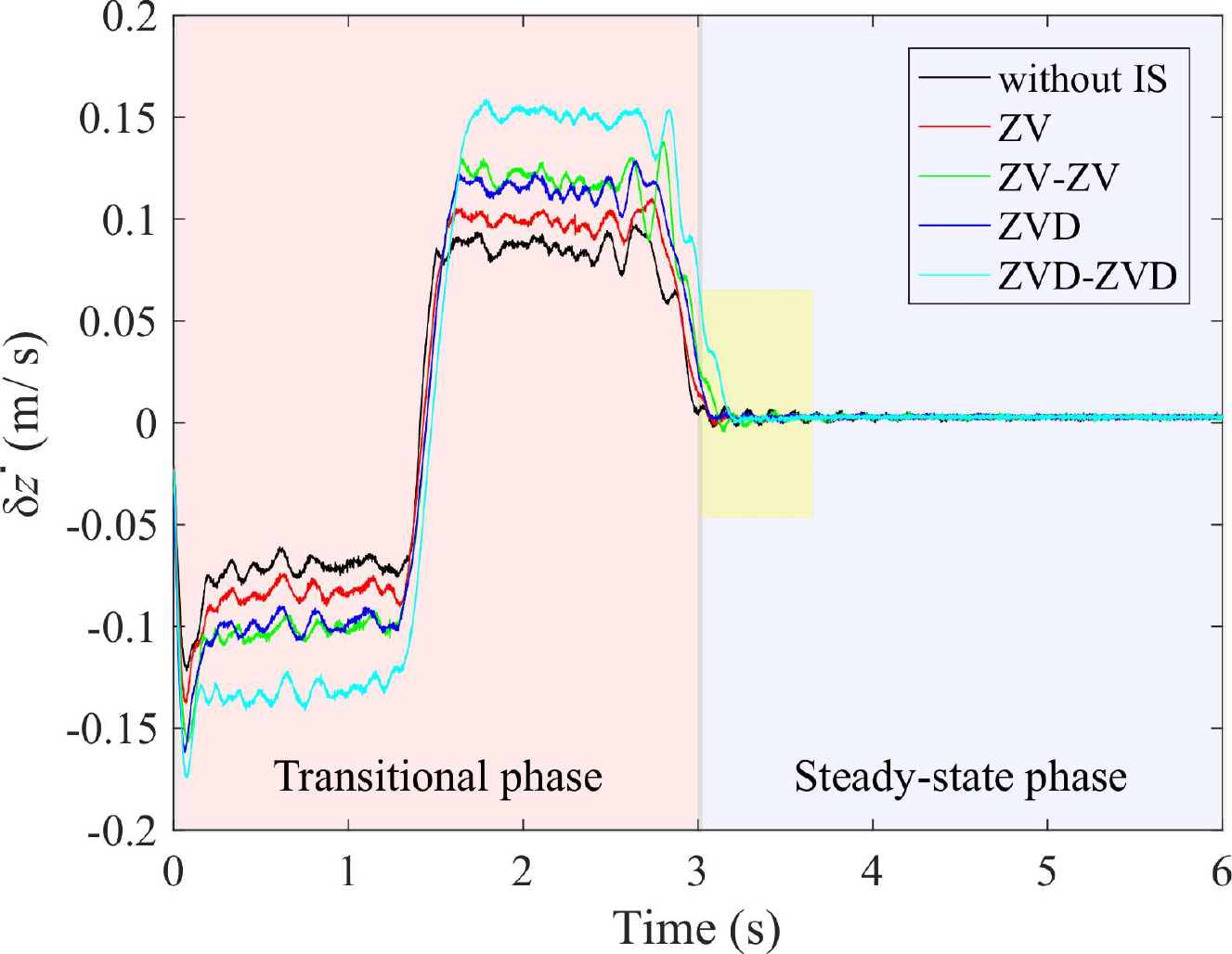}\label{fig:dzis}}\\
	\subfloat[]{\includegraphics[width=0.45\linewidth]{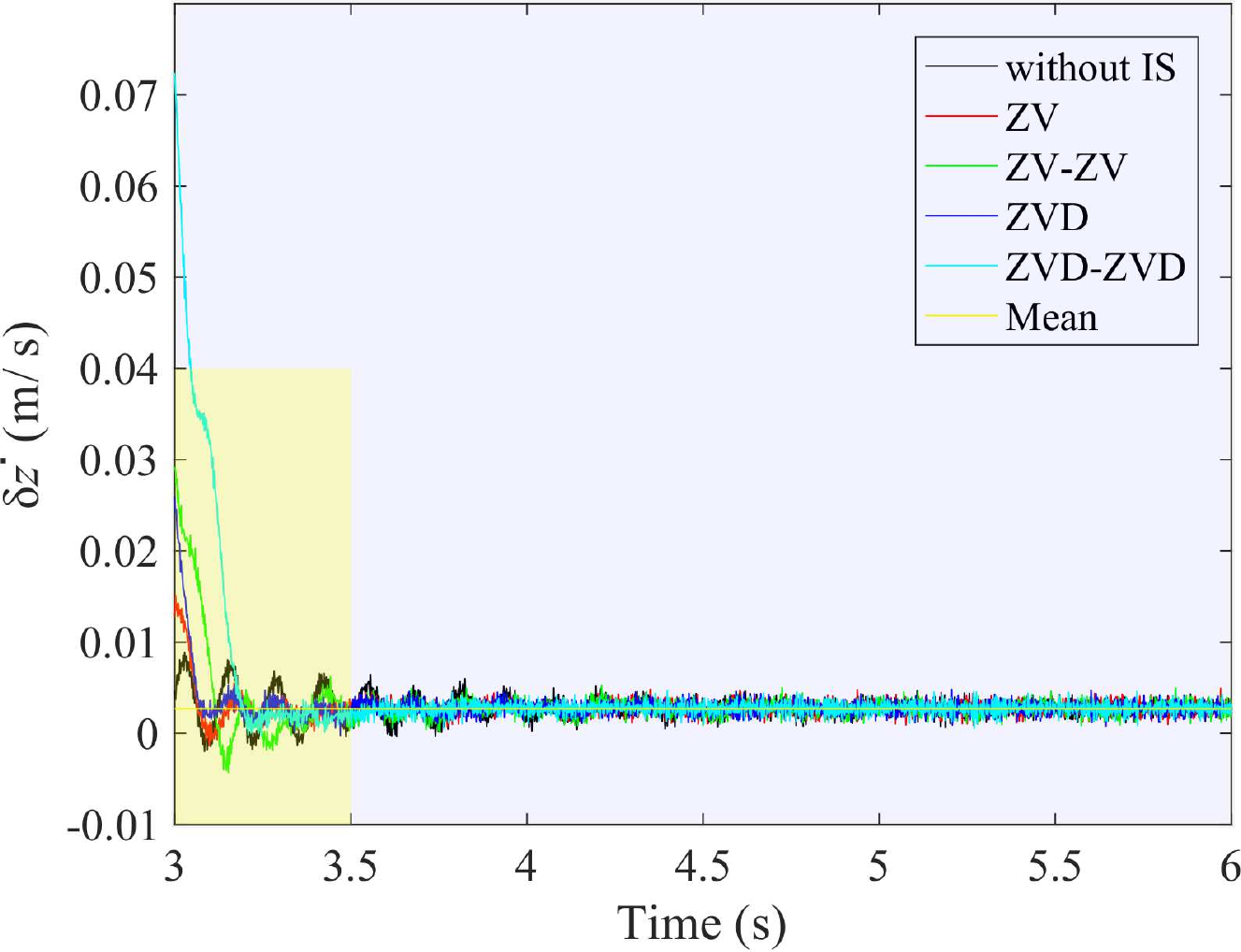}\label{fig:dzis_ss}}
	\subfloat[]{\includegraphics[width=0.45\linewidth]{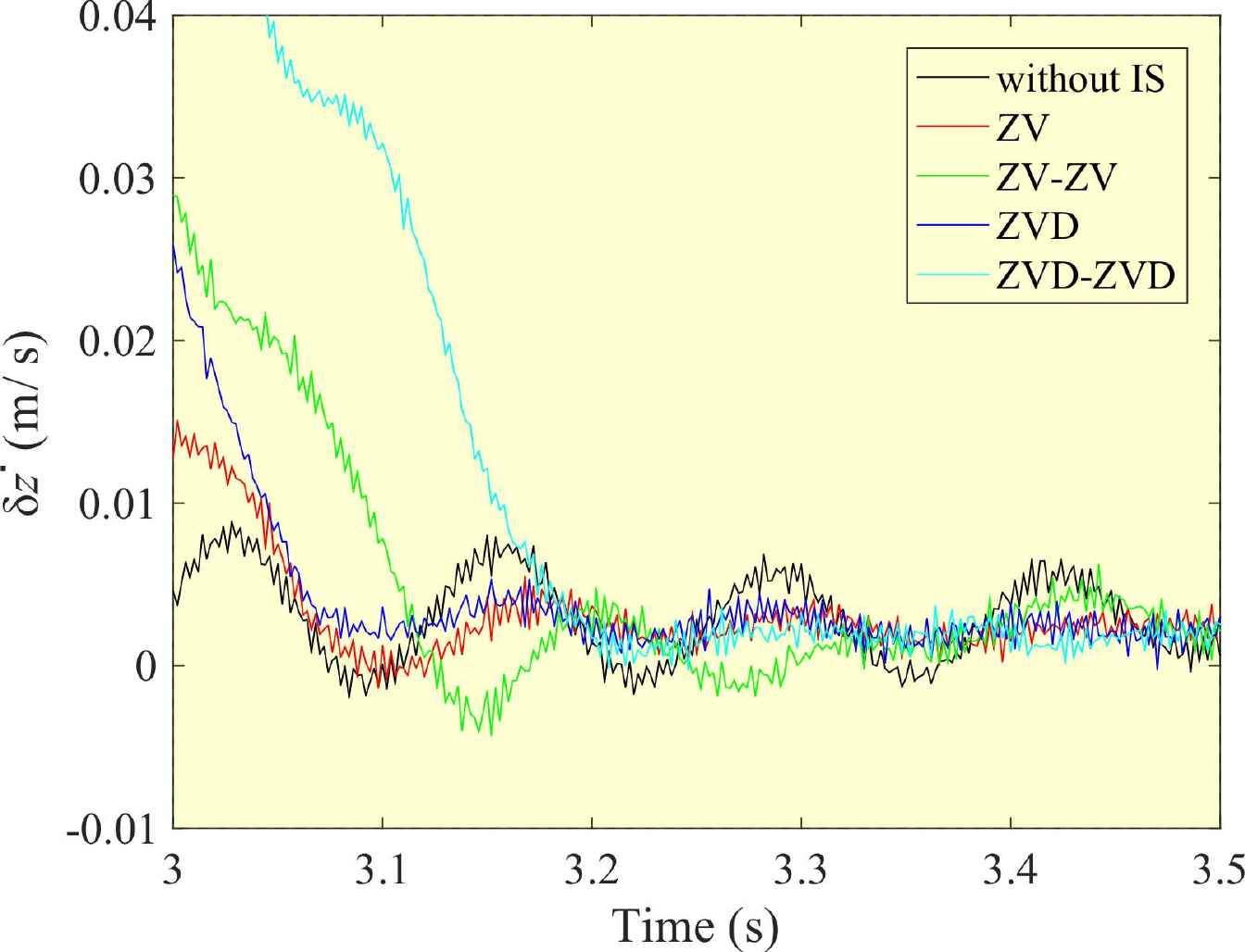} \label{fig:dzis_zoom}}
	\caption{Experimental results: Moving-platform (a)~Velocity error along z-axis with and without input-shaping (b)~During the steady-state phase (c)~Zoom on the moving-platform velocity error with and without input-shaping during the steady-state phase}
\end{figure}

Figures~\ref{fig:dzis} shows the velocity error $\delta \dot{z}$ of the moving-platform along z-axis, which is defined as the difference between the nominal velocity of the moving-platform and the measured one.
 The black (red, blue, green, cyan, resp.) plot depicts $\delta \dot{z}$  when the unshaped (ZV-shaped, ZVD-shaped, ZV-ZV-shaped, ZVD-ZVD-shaped, resp.) motion is used as a reference. 

To compare the results obtained with the input-shaping filters, we focus on the oscillations generated by discontinuities. Accordingly, a zoom is made at time range t~$\in~[3~~6]$~s, as the moving-platform is supposed to become motionless at time $t$~=~3~s and then residual vibrations appear. From Fig.~\ref{fig:dzis_zoom}, it appears that the ZVD-ZVD shaper is the fastest one in terms of residual vibration attenuation. 

\begin{figure}[!htb]
	\centering
	\includegraphics[width=0.5\linewidth]{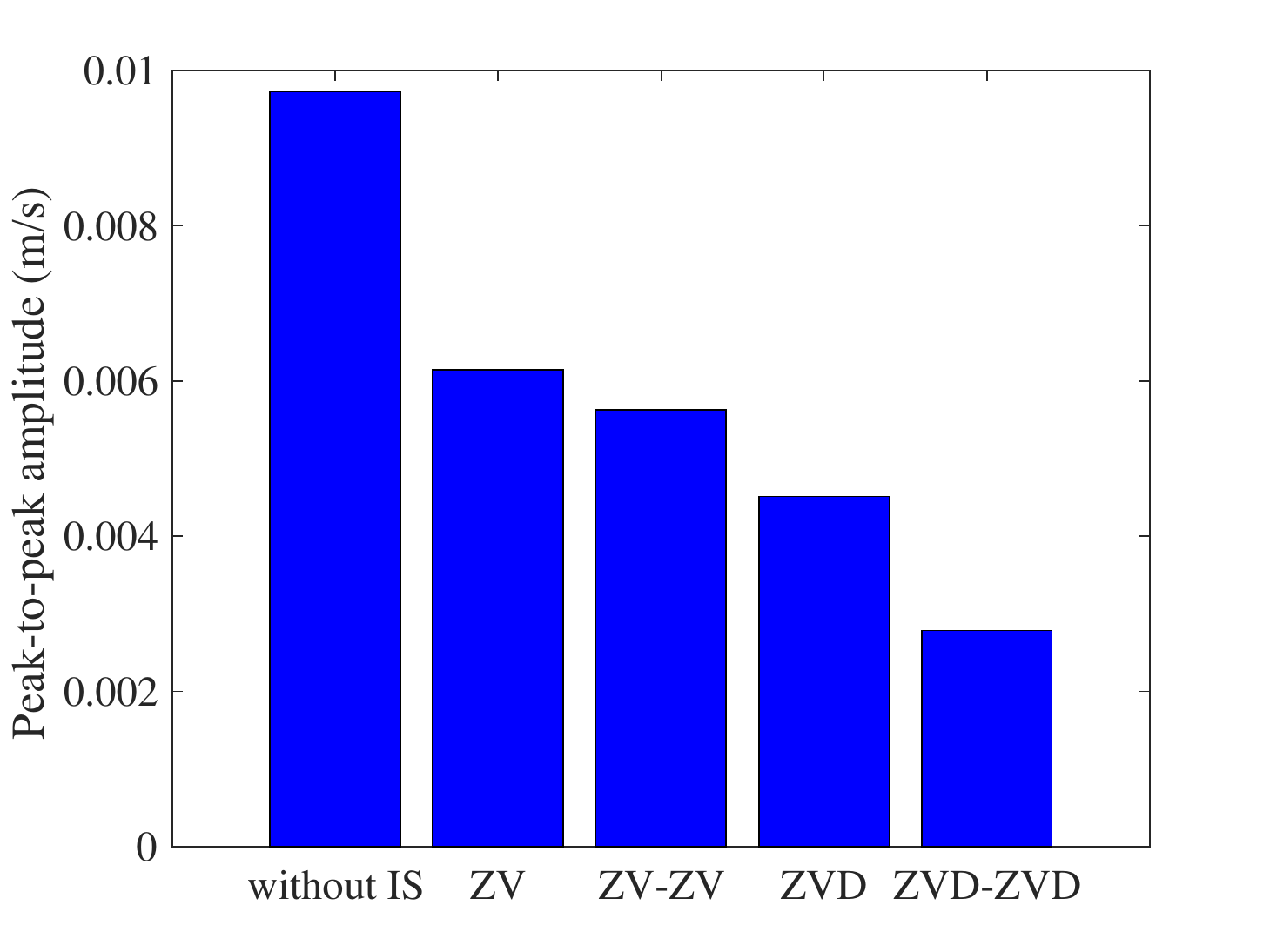}
	\caption{Bar chart of the first period Peak-to-Peak amplitude of $\delta \dot{z}$}
	\label{fig:barchartis}
\end{figure}

To better compare the performances of the different input-shapers, Fig.~\ref{fig:barchartis} represents a bar chart showing the first Peak-to-Peak amplitude of $\delta \dot{z}$. This period starts when the $\delta \dot{z}$ curve intersects with the line presenting the mean value of oscillations. 
When no input-shaper is used, the Peak-to-Peak amplitude of $\delta \dot{z}$ is equal to 0.0097~m/s. It is equal to 0.0061~m/s, 0.0045~m/s, 0.0056~m/s and to 0.0028~m/s when the ZV, ZVD, ZV-ZV and the ZVD-ZVD shapers are used, respectively. These results confirm that the ZVD and the ZVD-ZVD shapers lead to the most stable behavior of the moving-platform.
 
Contrary to the ZV-shaper, the ZVD-shaper is more robust to modeling errors. The convolution of two one-mode robust shapers results in a two-mode robust shaper dealing with two natural modes of the robot under study. That explains the fact that the ZVD-ZVD filter leads to better oscillation attenuations. The two-mode ZVD-ZVD input-shaping filter leads to the most robust control scheme and to the best one in terms of vibration rejection. Nevertheless, this robustness incurs in a time penalty (a delay of 0.23~s with ZVD-ZVD shaper vs. a delay of 0.144~s with ZV-ZV shaper) so that the non-robust ZV-shapers are faster (Fig.~\ref{fig:dzis}) than the robust ZVD-shapers.

\begin{figure}[!htb]
	\centering
	\subfloat[]{\includegraphics[height=0.35\linewidth]{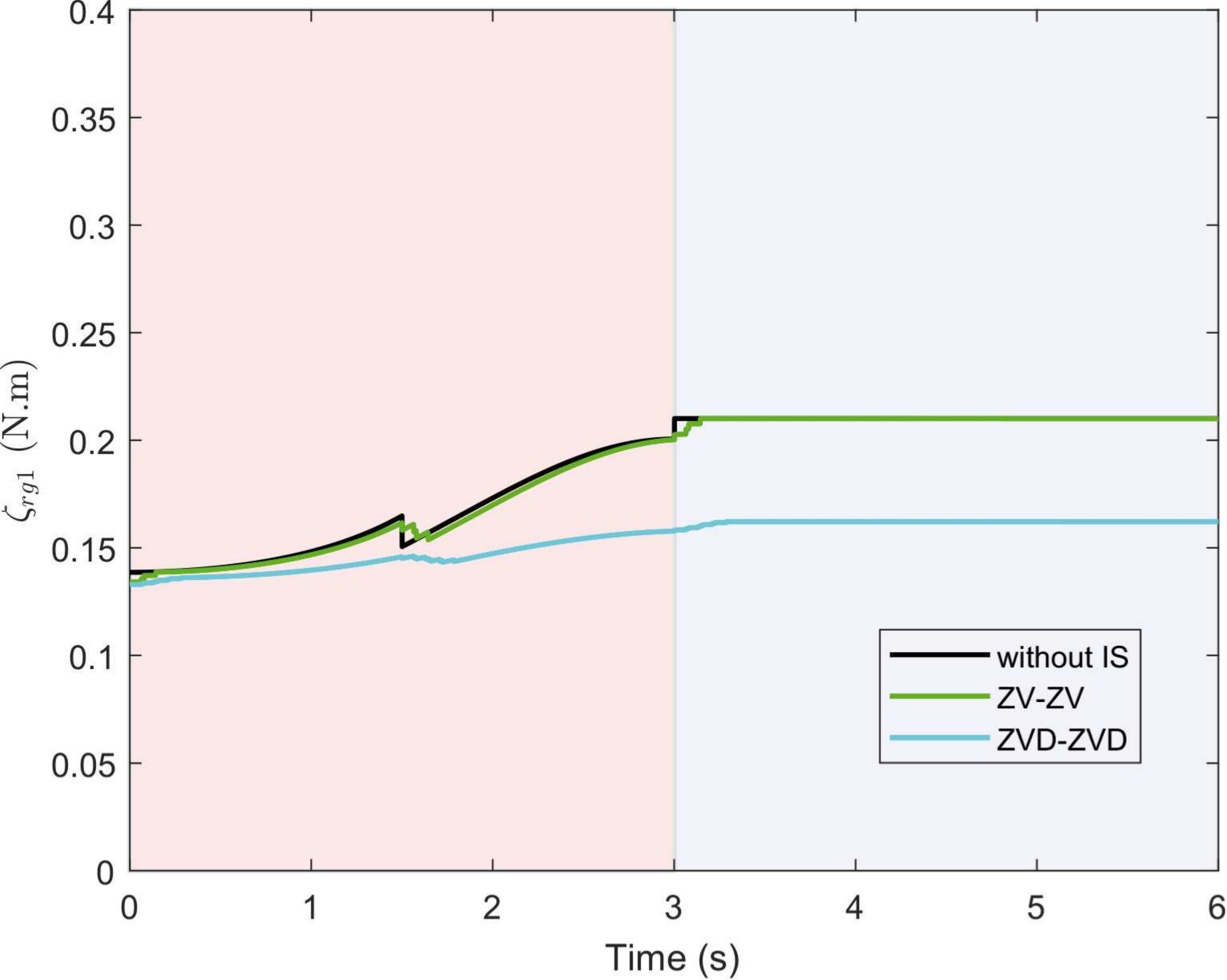}\label{fig:compforceis}}
	\subfloat[]{\includegraphics[height=0.35\linewidth]{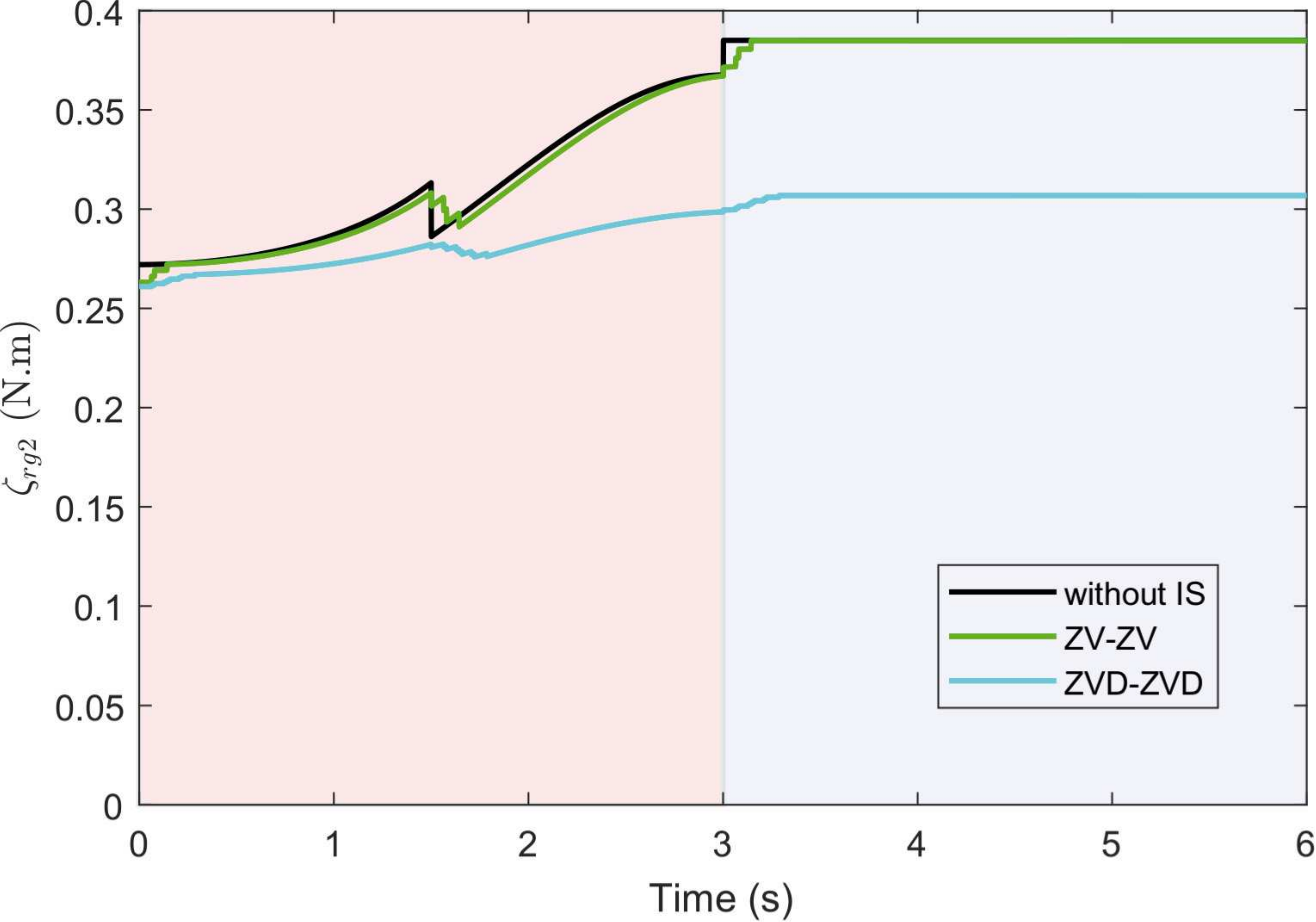}\label{fig:compforceis2}}\\
	\subfloat[]{\includegraphics[height=0.35\linewidth]{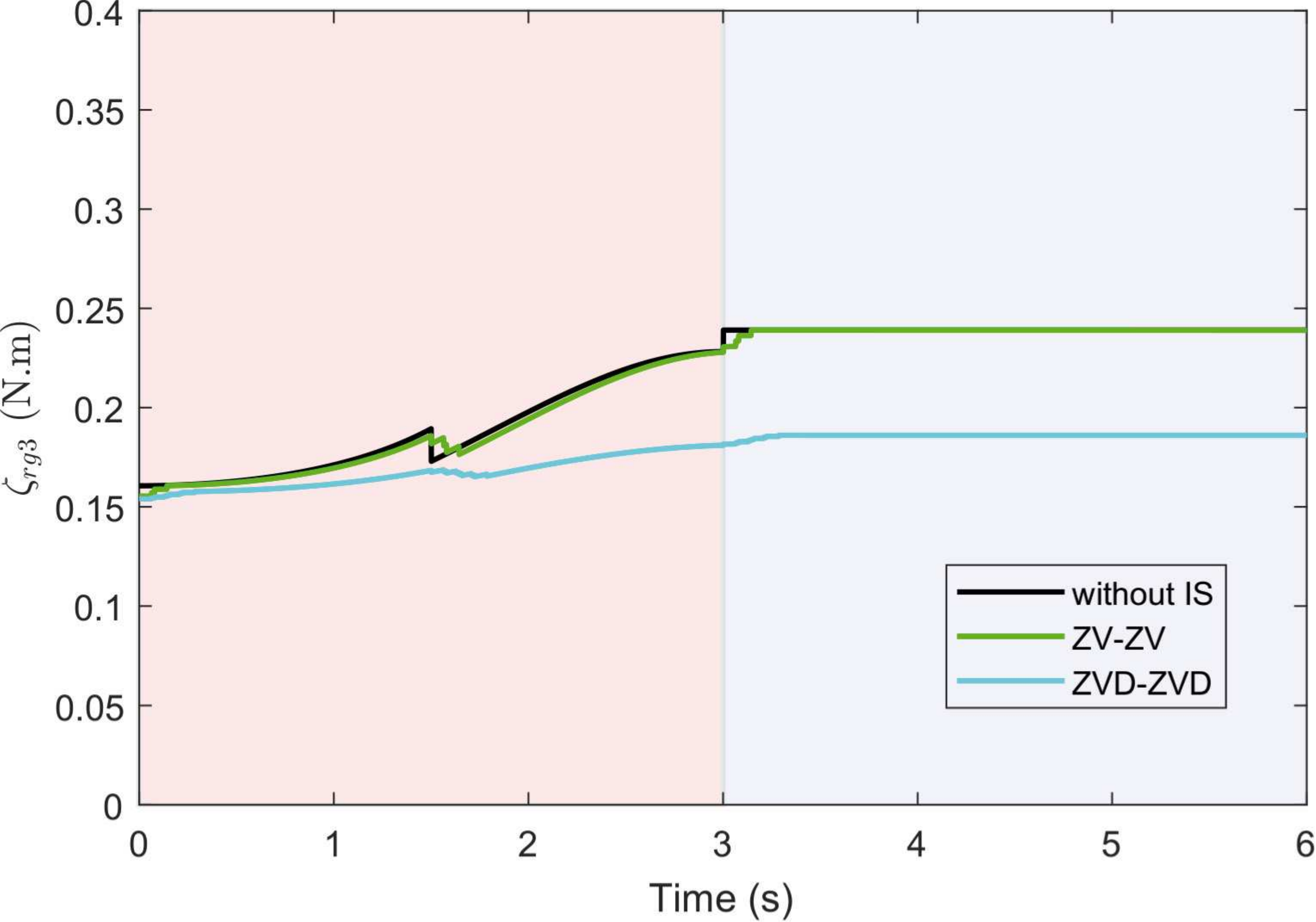} \label{fig:compforceis3}}
	\caption{Cable torque set-point without IS, with ZV-ZV shaper and with ZVD-ZVD shaper: (a)~cable~1, (b)~cable~2, (c)~cable~3}
	\label{fig:comp_force_inputshaping}
\end{figure}

Besides vibration reduction, the input-shaping reduces the control effort. Figure~\ref{fig:comp_force_inputshaping} shows cable torque set-point along the shaped trajectory with and without input-shaping. The torque set-point of the first (second, third, resp.) cable is equal to 0.138~N.m (0.272~N.m, 0.160~N.m, resp.) at start time when no input-shaping is applied. It is equal to 0.134~N.m (0.263~N.m, 0.155~N.m) when ZV-ZV shaping is applied, which presents a torque reduction of 2.89~\% (3.3\%, 3.12~\%, resp.). This value is equal to 0.133~N.m (0.261~N.m, 0.154~N.m, resp.) when ZVD-ZVD input-shaping is applied, which presents a torque reduction of 3.62~\% (4.04\%, 3.75~\%, resp.) with respect to the non modified torque. The torque set-point of the first (second, third, resp.) cable is equal to 0.210~N.m (0.385~N.m, 0.228~N.m, resp.) at the end of transitional phase when no input-shaping is applied. It is equal to 0.200~N.m (0.367~N.m, 0.226~N.m, resp) when ZV-ZV shaping is applied, which presents a torque reduction of 4.76~\% (4.67~\%, 0.88~\%, resp.). This value is equal to 0.156~N.m (0.298~N.m, 0.181~N.m, resp.) when ZVD-ZVD input-shaping is applied, which presents a torque reduction of 25.71~\% (22.59\%, 20.61~\%, resp.) with respect to the non modified torque.

\subsection{Discussion}
It is clear that the use of input-shapers leads to better attenuation of residual vibrations, especially while using robust input-shapers. However, these vibrations are not totally attenuated.

It is noteworthy that an important part of damping comes from the actuators. The non consideration of these damping parameters may degrade partially the effectiveness of input-shaping filters.

Note that the choice of constant modal parameters for the input-shapers in this paper was validated by a robustness analysis. This is not true for all CDPR geometric configurations and/or payloads. The degradation of input-shaper performance can be caused by mechanical parameters and payload uncertainties and/or variations due to evolution over the workspace~\cite{solatges2017adaptive}. In this case, adaptive input-shapers as proposed in~\cite{tzes1993adaptive, khorrami1995experimental, kojima2007adaptive} should be used in order to reduce uncertainties in modal parameters that lead to residual vibrations.

\begin{figure}[!htb]
	\centering
	\includegraphics[width=0.45\linewidth]{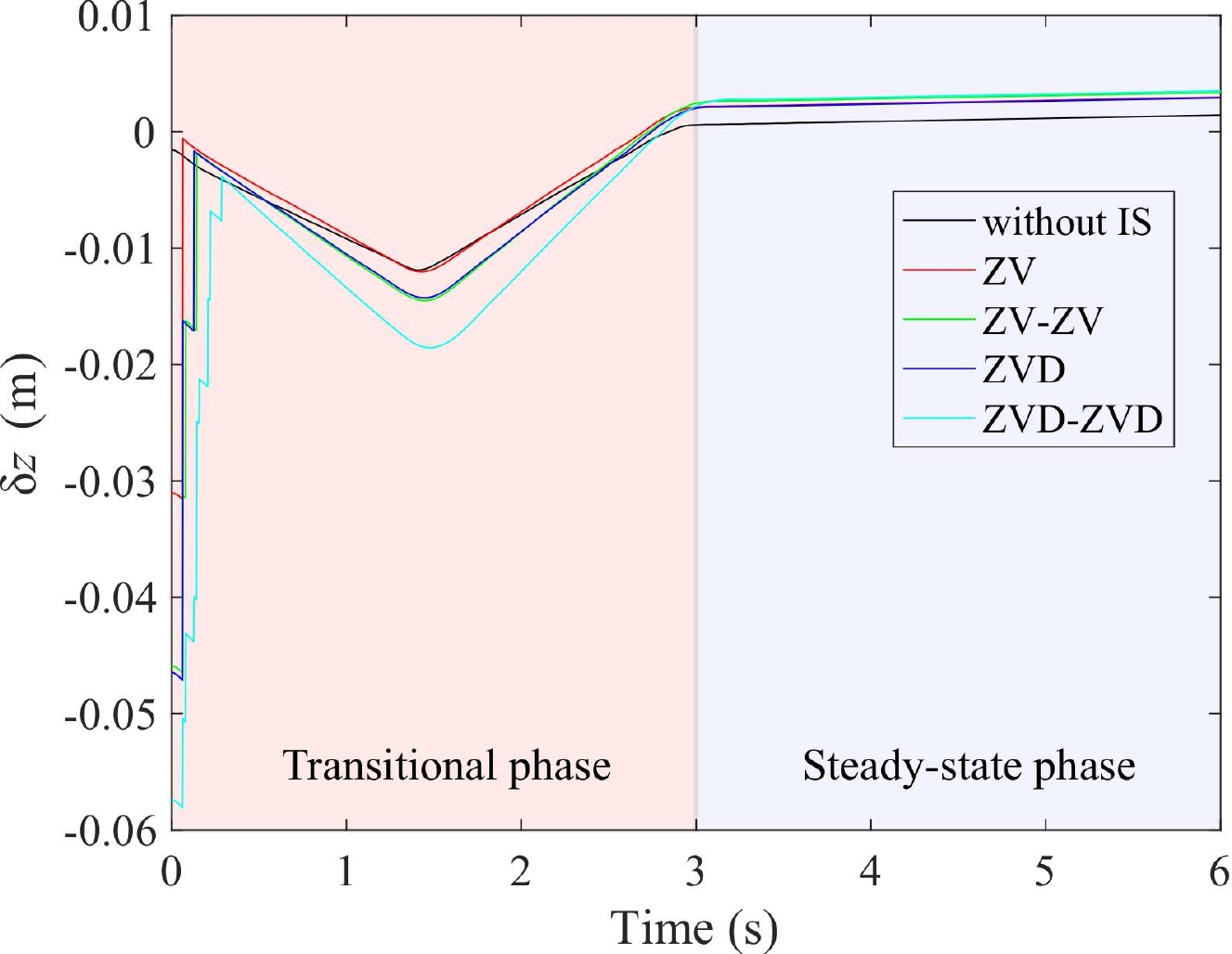}
	\caption{Experimental results: Moving-platform position error along z-axis with and without input-shaping}
	\label{fig:erdepz}
\end{figure}

As shown in Fig.~\ref{fig:erdepz}, the use of input-shapers into the feed-forward control degrades the moving-platform trajectory tracking due to the delays and impulses applied to the original control signal.
To deal with this issue, it would be better to combine the control scheme proposed in this paper with the work presented in \cite{baklouti2019vibration}. This latter focuses on the compensation of the moving-platform deflection coming not only from the elastic behavior of cables but also from the moving-platform dynamics. The efficiency of such a control scheme in terms of trajectory tracking improvement has been experimentally validated for a non redundant and suspended CDPR \cite{baklouti2019vibration}. 

\section{Conclusion}
\label{sec:conclusion}
This paper proposed a frequency-dependent method to attenuate the unwanted vibrations of Cable-Driven Parallel Robots~(CDPRs). This method deals with the integration of input-shaping filters into the closed-loop feed-forward control. Two classes of input-shapers were proposed: single-mode and multi-mode input-shapers. These filters re-design the input signal to cancel residual vibrations. A new type of workspace related to the shaper sensitivity to the variations in natural frequencies was defined. This workspace aims to verify the shaper performance along a prescribed trajectory based on its starting point. The comparison between the velocity errors obtained through experimentations when using unshaped input signal or shaped ones as control references shows meaningful differences. Accuracy improvement with respect to the unshaped control in terms of Peak-to-Peak amplitude of velocity error achieves 53~\% while using the ZVD input-shaper and 72~\% while using the ZVD-ZVD input-shaper. This percentage is equal to 36~\% with a ZV input-shaper and to 42~\% with a ZV-ZV input-shaper. Experimental results confirm that the integration of input-shaping filters into the closed-loop feed-forward control scheme is relevant to attenuate residual vibrations of a non-redundant CDPR. Further experiments will be performed while compensating cable elongations and considering redundantly actuated CDPRs later on. 
Integrating an adaptive input-shaper into the feed-forward control should further improve the overall CDPR performance. In order to improve the performance of multi-mode input-shaping control schemes, an on-line estimation of CDPR damped natural frequencies, with respect to CDPR geometry, cable tensions and the moving-platform trajectory, will allow the user to tune input-shaping filters as a function of those estimated along a prescribed trajectory of the moving-platform and to reduce the moving-platform pose stabilization time.

\section{Nomenclature}

\begin{description}
	\item[\rm{DOF}]: Degree-Of-Freedom.
	\item[\rm{CDPR}]: Cable-Driven Parallel Robot.
	\item[\rm{ZV}]: Zero-Vibration.
	\item[\rm{ZVD}]: Zero-Vibration-Derivative.
	\item[$\displaystyle{\mathcal{F}_b=\{O,~x_b,~y_b,~z_b\}}$]: Base frame. 
	\item[$\displaystyle{\mathcal{F}_p=\{P,~x_p,~y_p,~z_p\}}$ ]: Moving-platform frame.
	\item[$n$]: Number of cables.
	\item[$m$]: Total number of DOF.
	\item [${K}_{p}$]: Proportional gain of PID controller. 
	\item [${K}_{i}$]: Integral gain of PID controller.
	\item [${K}_{d}$]: Derivative gain of PID controller.
	\item [$f$]: Natural frequency. 
	\item [$\zeta$]: Damping ratio.
 	\item[$\mathbf{a}_i$]: Cartesian coordinate vector of anchor points $A_i$ expressed in $\mathcal{F}_p$.
	\item[$\mathbf{b}_i$]: Cartesian coordinate vector of exit points $B_i$ expressed in $\mathcal{F}_b$. 
	\item[$\mathbf{x}$]: Moving-platform pose vector expressed in $\mathcal{F}_b$.
	\item[$\mathbf{p}$]: Moving-platform position vector.
	\item[$\mathbf{o}$]: Moving-platform orientation vector.
	\item[$\mathbf{t}$]: Moving-platform twist.
	\item[$\boldsymbol{\Gamma}$]: Motor torque set-point vector.
	\item{$\boldsymbol{\tau}$}: Cable tension vector.
	\item[$\mathbf{q}$]: Angular displacement vector.
	\item[$\mathbf{l}$]: Cable length vector.
	\item[$\boldsymbol{\chi}$]: Diagonal matrix containing the winding ratio of the winches.
	
\end{description}

\printnomenclature

\bibliographystyle{asmems4}

\bibliography{asme2e}

\end{document}